\documentclass{article}
\usepackage{algorithm}
\usepackage[noend]{algpseudocode}
\usepackage{xcolor}
\usepackage{setspace}
\usepackage{algpseudocode}
\usepackage{amsmath}
\usepackage{subcaption}
\usepackage{amsmath}
\usepackage{amssymb}
\usepackage{caption}
\usepackage{algorithm}
\usepackage{algpseudocode}
\usepackage{amsmath}
\usepackage[export]{adjustbox}
\usepackage[utf8]{inputenc}
\usepackage{array}
\usepackage{color}
\usepackage{xcolor}
\usepackage{hyperref}

\usepackage{arxiv}

\usepackage[utf8]{inputenc} 
\usepackage[T1]{fontenc}    
\usepackage{hyperref}       
\usepackage{url}            
\usepackage{booktabs}       
\usepackage{amsfonts}       
\usepackage{nicefrac}       
\usepackage{microtype}      
\usepackage{lipsum}
\usepackage{graphicx}
\graphicspath{ {./images/} }

\title{Explainable Differential Privacy-Hyperdimensional Computing for Balancing Privacy and Transparency in Additive Manufacturing Monitoring}

\author{
Fardin Jalil Piran \\
  School of Mechanical, Aerospace, and Manufacturing Engineering\\
  University of Connecticut\\
  Storrs, CT 06269 \\
  \texttt{fardin.jalil\_piran@uconn.edu} \\
   \And
 Prathyush P. Poduval \\
  Department of Computer Science\\
  University of California Irvine\\
  Irvine, CA 92697\\
  \texttt{ppoduval@uci.edu} \\
  \And
   Hamza Errahmouni Barkam \\
  Department of Computer Science\\
  University of California Irvine\\
  Irvine, CA 92697\\
  \texttt{herrahmo@uci.edu} \\
  \And
 Mohsen Imani \\
  Department of Computer Science\\
  University of California Irvine\\
  Irvine, CA 92697\\
  \texttt{m.imani@uci.edu} \\
  \And
Farhad Imani\\
  School of Mechanical, Aerospace, and Manufacturing Engineering\\
  University of Connecticut\\
  Storrs, CT 06269 \\
  \texttt{farhad.imani@uconn.edu} 
}

\begin{document}
\maketitle
\begin{abstract}
Machine Learning (ML) models integrated with in-situ sensing offer transformative solutions for defect detection in Additive Manufacturing (AM), but this integration brings critical challenges in safeguarding sensitive data, such as part designs and material compositions. Differential Privacy (DP), which introduces mathematically controlled noise, provides a balance between data utility and privacy. However, black-box Artificial Intelligence (AI) models often obscure how this noise impacts model accuracy, complicating the optimization of privacy–accuracy trade-offs. This study introduces the Differential Privacy-Hyperdimensional Computing (DP-HD) framework, a novel approach combining Explainable AI (XAI) and vector symbolic paradigms to quantify and predict noise effects on accuracy using a Signal-to-Noise Ratio (SNR) metric. DP-HD enables precise tuning of DP noise levels, ensuring an optimal balance between privacy and performance. The framework has been validated using real-world AM data, demonstrating its applicability to industrial environments. Experimental results demonstrate DP-HD’s capability to achieve state-of the-art accuracy (94.43\%) with robust privacy protections in anomaly detection for AM, even under significant noise conditions. Beyond AM, DP-HD holds substantial promise for broader applications in privacy-sensitive domains such as healthcare, financial services, and government data management, where securing sensitive data while maintaining high ML performance is paramount.
\end{abstract}


\textbf{DOI:} \href{https://doi.org/10.1016/j.engappai.2025.110282}{https://doi.org/10.1016/j.engappai.2025.110282}
\section{Introduction}
Additive Manufacturing (AM) represents a transformative approach to production, enabling the fabrication of complex geometries through the sequential addition of material layers as directed by computer-aided design data~\cite{frazier2014metal}. Despite the revolutionary potential, AM faces significant challenges in ensuring part quality. Approximately 10\% of AM-produced parts fail due to various defects such as delamination, inclusion of foreign materials, porosity, incomplete fusion, and layer misalignment~\cite{leach2011calibration,mandache2019overview}. These defects compromise both the dimensional accuracy and structural integrity of the manufactured parts, often necessitating extensive and costly post-processing treatments, including hot isostatic pressing or precision machining to meet stringent quality standards~\cite{mandache2019overview, chen2023brain}. While new standards, inspections, and rectifications are being developed for post-build process qualification and part certification, experiments and simulation studies are being conducted to gain insight into the complex physics of AM processes, many of these efforts remain disconnected and inefficient~\cite{imani2019joint,yao2018multifractal}. 

The integration of in-situ sensing and monitoring significantly advances agile defect detection, drastically reducing production time and costs by facilitating real-time adjustments to process parameters, thus enhancing the repeatability and reliability of AM processes~\cite{shevchik2019deep}. Machine Learning (ML) models have significantly enhanced AM by leveraging data to automatically diagnose printing status~\cite{wu2017real}, identify failure modes~\cite{he2018intelligent}, calibrate systems~\cite{rescsanski2022heterogeneous}, assess melting conditions~\cite{ye2018defect, grasso2018situ}, detect porosity~\cite{imani2018process}, predict tensile properties~\cite{okaro2018automatic} and surface roughness~\cite{li2019prediction}, as well as to perform feedback control~\cite{clijsters2014situ, craeghs2010feedback}, model-based feed-forward control~\cite{yeung2019part, khairallah2020controlling}, and distortion compensation~\cite{kumar2019distortion}. However, significant privacy risks arise due to the manufacturing sector's vulnerability to cyberattacks. In 2022, the manufacturing sector experienced the highest rate of cyber incidents at 23.2\%, surpassing even finance and insurance~\cite{IBM2022}. These cyberattacks, including ransomware and data breaches, have serious financial impacts, with average breach costs exceeding \$4.35 million. Small and medium-sized manufacturing firms are particularly vulnerable; in 2017, 61\% of these firms reported cyberattacks, with median breach costs over \$60,000~\cite{Ponemon2019}. Implementing in-situ sensing techniques exposes sensitive operational data, increasing the risk of unauthorized access. Model inversion attacks are particularly concerning, as adversaries exploit ML model vulnerabilities to extract confidential data about product specifications, process parameters, and machinery used in predictive analytics~\cite{wang2015regression, hu2020privacy}. 

To mitigate privacy concerns in in-situ sensing, several privacy-preserving techniques are employed, including cryptographic methods, data anonymization, and Differential Privacy (DP). Cryptographic methods safeguard identities and manage data access, ensuring the confidentiality and integrity of sensitive information~\cite{cambou2020securing}. However, substantial computational power and higher bandwidth are often required, hindering the real-time processing essential for in-situ monitoring. Data anonymization involves removing identifying information from datasets prior to analysis~\cite{neves2023data}, aiding in the management of complex, high-dimensional data, and reducing the risk of data breaches. Nevertheless, this technique results in the loss of valuable data granularity and does not fully guarantee privacy. The potential for re-identification remains high, as demonstrated by instances where anonymized Netflix data was re-identified by correlating it with external data~\cite{narayanan2008robust}.

\textcolor{black}{DP has emerged as an effective method for ensuring confidentiality in AM by strategically adding controlled noise to data to protect data privacy~\cite{dwork2006calibrating, dwork2014algorithmic}. DP has been successfully applied in various industries, notably healthcare, to protect sensitive information against model inversion attacks~\cite{fredrikson2014privacy, krall2020mosaic}. However, employing DP requires careful management of the trade-off between enhanced privacy and the performance of ML models. Higher levels of noise can strengthen privacy protections but may also degrade algorithm precision and functionality. Thus, finding an optimal balance that ensures robust data protection while retaining process utility and efficiency is essential. A significant gap in existing approaches is compounded by the "black-box" nature of ML models, which obscures how DP noise impacts model accuracy, making it difficult to predict or fine-tune the privacy-accuracy balance. This lack of transparency and predictability in black-box models poses considerable limitations in optimizing DP applications effectively. Consequently, there is a critical need for advancements in Explainable Artificial Intelligence (XAI) that make the effects of DP noise on model accuracy more interpretable and quantifiable, particularly in privacy-sensitive AM applications~\cite{lubell2022protecting}.}

\textcolor{black}{To effectively anticipate the influence of DP noise on model accuracy, this paper introduces the Differential Privacy-Hyperdimensional Computing (DP-HD) framework. This explainable approach is tailored for in-process monitoring, featuring rapid query response capabilities, making it ideal for applications requiring swift evaluation while preserving both privacy and performance fidelity. The strength of DP-HD lies in its innovative data processing method, which utilizes high-dimensional vector spaces for computation. This design not only enables efficient analysis but also provides a way to define privacy levels precisely and predict model accuracy before adding noise, aided by the introduction of a Signal-to-Noise Ratio (SNR) metric to measure the balance between data contribution and DP noise. A case study using real-world AM data, representative of typical industrial conditions, demonstrates the practical benefits of DP-HD, showing its ability to safeguard sensitive data while maintaining high accuracy in defect detection. Beyond AM, this framework offers potential for other privacy-sensitive industries, such as aerospace and healthcare, where protecting proprietary data is critical. The primary contributions of this work are as follows:}
\begin{enumerate}
    \item \textcolor{black}{We propose an explainable DP-HD framework to enhance transparency in interactions between ML models and DP. The framework introduces a method for determining the optimal Standard Deviation (SD) for generating random basis vectors, aided by the use of an SNR metric to achieve an ideal balance between privacy protection and performance accuracy.}
    \item \textcolor{black}{We evaluate the impact of noise on the framework’s memorization capacity, demonstrating how the SNR metric enables DP-HD to accurately predict noise effects on model accuracy across various privacy levels, a critical capability for privacy-sensitive applications.}
    \item \textcolor{black}{We present a novel inference privacy technique that reduces the risk of decoding query hypervectors by removing low-variance dimensions. This technique significantly enhances the security of DP-HD, making it robust and scalable for real-world industrial applications.}
\end{enumerate}

The structure of this paper is organized as follows: Section~\ref{sec:Research Background} discusses previous works related to in-situ sensing in AM and the preservation of privacy within these systems. Section~\ref{sec:Research Methodology} details the proposed methodology. Section~\ref{sec:Experimental Design} outlines the experimental setup and describes the design used in our case study, utilizing real-world AM data. Section~\ref{sec:Experimental Results} discusses the findings from the experiments. Section~\ref{sec:Discussion} provides an analysis of the implications of our results, examining the strengths, limitations, and broader impact of the proposed framework. Finally, Section~\ref{sec:Conclusions} concludes the paper by highlighting the limitations of current optimization methods for parameter selection in AM and suggesting potential avenues for future research.

\section{Research Background}
\label{sec:Research Background}

\subsection{In-situ Sensing and Monitoring}

AM employs various sensing techniques for defect detection, including laser scanning~\cite{faes2016process}, thermographic methods~\cite{krauss2012thermography}, radiographic~\cite{shevchik2020supervised}, ultrasound~\cite{koskelo2016scanning}, electromagnetic~\cite{lou2018internal}, and acoustic emission~\cite{wang2008online}. Visual imaging methods, in particular, are prized for their ability to automate the detection and evaluation of defects, making them critical for in-situ quality assurance in AM~\cite{han2020defect}. Commonly used equipment includes visible light cameras with specialized filters and fast shutters, or infrared cameras to monitor the AM process~\cite{abouelnour2022situ}. For example, Barua et al. utilized digital cameras with macro lenses to observe the laser metal deposition melt pool, creating a temperature map from Red, Green, and Blue (RGB) data through linear regression, which helps in identifying defects by highlighting variations in thermal conductivity~\cite{barua2014vision}.

Visual imaging techniques also play a significant role in enhancing defect identification. By varying imaging angles, defects can be revealed through contrast changes~\cite{shen2012bearing}, while structured light projection on surfaces is effective in detecting specular reflections. Techniques such as extended Gaussian images can represent 3D surface shapes and orientation histograms, facilitating shape classification without the need for feature localization and matching. Despite the improved detection precision offered by structured lighting sources like stripes, they require extensive computational power, which limits their practicality for real-time AM monitoring~\cite{perard1997three}. To enhance computational efficiency, Aluze et al. proposed alternating between illuminated and non-illuminated stripes to improve the contrast between defective and intact areas. However, this method's effectiveness may be limited on surfaces where light paths do not reflect well off concave geometries~\cite{aluze2002vision}.

Furthermore, inline coherent imaging, which uses low coherence interferometry, creates detailed 3D models similar to spectral-domain optical coherence tomography and is particularly effective for analyzing melt pool geometries. Madrigal et al. employed distorted light patterns and a feature descriptor known as the model points feature histogram for model generation, which were then used as inputs for support vector machines in classification tasks~\cite{madrigal2017method}. Additionally, Holzmond et al. used two digital cameras placed at opposite angles to create 3D point clouds representing surface geometry~\cite{holzmond2017situ}. Defect detection involved comparing vertical discrepancies (Z-differences) between points in the point cloud and a reference 3D mesh. The iterative closest point algorithm aligned these points to calculate necessary spatial transformations. Defect classification was based on a set threshold for Z-differences, although further documentation is needed to validate the method's accuracy.

While some defect detection methods for AM processes support in-situ applications, many are unable to effectively identify small defects or lack reliable accuracy metrics~\cite{li2021geometrical}. ML models ensure product quality and enhance the efficiency of the printing process in AM~\cite{fu2022machine}, as they excel at extracting relevant features from complex datasets. ML has been widely applied in manufacturing contexts~\cite{razvi2019review}, highlighting its capability for defect detection in real-time, examining powder spreading, and optimizing process parameters~\cite{wang2020machine}. Recent studies have successfully employed ML in various AM processes. For instance, defect detection in Laser Powder Bed Fusion (LPBF) has been improved using diverse ML and deep learning techniques, achieving accuracies above 80\%~\cite{scime2018anomaly,gobert2018application}. Moreover, ML and deep learning methods have been applied to quality monitoring in selective laser melting~\cite{caggiano2019machine} and fused filament fabrication~\cite{narayanan2019support}. Convolutional Neural Networks (CNNs) have also been proposed to assess the quality of the powder bed fusion process by utilizing high-speed cameras for image data acquisition~\cite{zhang2018extraction}.

State-of-the-art ML models for image classification include ResNet50, GoogLeNet, AlexNet, DenseNet201, and EfficientNet B2. ResNet50 employs deep residual learning, allowing for the training of deeper networks through residual blocks with skip connections, which improve gradient flow and mitigate the vanishing gradient problem, achieving excellent results on complex tasks~\cite{he2016deep}. GoogLeNet, also known as Inception v1, uses inception modules to perform convolutions of various sizes alongside max pooling, optimizing computational resources and enhancing network depth and width without significant computational overhead~\cite{szegedy2015going}. AlexNet revitalized interest in CNNs with its deep convolutional layers, max-pooling, Rectified Linear Unit (ReLU) activation functions, dropout, and fully connected layers, drastically reducing error rates in image classification tasks~\cite{krizhevsky2012imagenet}. DenseNet201 connects each layer to every other layer in a feed-forward manner, promoting feature reuse and reducing the number of parameters through dense connectivity, which supports effective feature propagation and efficiency in preserving feature information~\cite{huang2017densely}. EfficientNet B2 builds on the base EfficientNet architecture by applying a compound scaling coefficient, balancing network depth, width, and resolution to enhance performance while being resource-efficient, making it ideal for high accuracy in constrained environments~\cite{tan2019efficientnet}.

Although ML models have demonstrated considerable promise for defect detection in AM, they remain vulnerable to model inversion attacks, which allow attackers to extract confidential information from the training data. These attacks leverage background information and the ML models themselves to reveal sensitive details, such as design files, material properties, production parameters, and operational data. Consequently, implementing privacy-preserving methods is crucial to protect this sensitive information and ensure the security of the manufacturing process.

\subsection{Privacy-Preserving in In-situ Sensing}

Incorporating noise into algorithms, as originally proposed by Dwork, is essential for privacy. This technique ensures that databases, even if they differ by just one record, remain indistinguishable when processed by the same algorithm, thus greatly enhancing privacy protection~\cite{dwork2014algorithmic}. The intentional addition of noise effectively obscures data, preventing the leakage of sensitive information and guarding against model inversion attacks. By embedding DP into AM processes, manufacturers can protect confidential data while allowing for secure and privacy-aware applications of ML in monitoring and analysis~\cite{hu2020privacy,lee2023privacy}. As AM technology progresses, incorporating DP and other privacy-preserving strategies will be essential in overcoming the sector's cybersecurity issues. These methods not only ensure the privacy of individual data but also provide a safeguard against advanced cyber threats, maintaining the security and integrity of the manufacturing process. However, implementing these methods necessitates a delicate balance between increasing privacy and maintaining algorithmic performance. While higher noise levels enhance privacy, they can also diminish the precision and efficiency of the algorithm. Therefore, it is crucial to establish an optimal balance that provides robust data protection while preserving the process's utility and effectiveness.

Several techniques have been developed to implement privacy-preserving ML algorithms through DP in AM. Hu et al. proposed optimizing the selection of DP mechanisms and privacy parameters to balance model utility and robustness against attacks~\cite{hu2020privacy}. Lee et al. introduced the mosaic neuron perturbation method, which perturbs neural network training by injecting carefully designed noise to ensure DP~\cite{lee2023privacy}. This method uses two control parameters to adjust the privacy level and the perturbation ratio between sensitive and non-sensitive attributes, reducing the risk of inversion attacks on sensitive features while retaining the model's predictive power.

\textcolor{black}{The state-of-the-art approach for applying DP to ML models is Differentially Private Stochastic Gradient Descent (DP-SGD), a variation of the standard Stochastic Gradient Descent (SGD) algorithm. DP-SGD enhances privacy by introducing precisely calibrated noise to the gradients during the training process. This technique minimizes the impact of individual data entries by clipping gradients and adding noise, thereby obscuring their specific contributions to the final model parameters. DP-SGD provides measurable privacy assurances, essential in fields where data confidentiality is paramount~\cite{abadi2016deep}.}

\begin{table}[h!]
\centering
\color{black}
\renewcommand{\arraystretch}{1}
\caption{\textcolor{black}{Summary of XAI Techniques: Advantages and Limitations}}
\begin{tabular}{p{4.8cm} p{4.8 cm} p{5.2 cm}}
\hline
\textbf{XAI Technique} & \textbf{Advantages} & \textbf{Limitations} \\
\hline
Attribution Methods~\cite{fan2021interpretability} & Provide clear relevance of input features to model decisions; easy to interpret as post-hoc explanations & Depend on model internals, making them less suitable for black-box models \vspace{2mm} \\

Saliency Methods (e.g., InputXGradient~\cite{shrikumar2016not}, GuidedBackpropagation~\cite{springenberg2014striving}, IntegratedGradients~\cite{sundararajan2017axiomatic}) & Efficiently highlight relevant features in neural networks using gradient-based approaches & Require access to network internals; limited in handling black-box scenarios \vspace{2mm}\\

Perturbation-based Methods (e.g., Occlusion~\cite{zeiler2014visualizing}) & Model-agnostic; can be applied to any architecture & Resource-intensive as they involve systematic input modifications; limited interpretability in complex models \vspace{2mm}\\

Surrogate Models (e.g., LIME~\cite{ribeiro2016should}, SHAP~\cite{lundberg2017unified}) & Provide local explanations; SHAP offers mathematical grounding through game theory & Computationally expensive, especially with high-dimensional data; may provide overly simplified explanations in complex models \vspace{2mm}\\

Prototype~\cite{li2018deep}, Patch~\cite{chen2019looks}, and Concept-based Methods~\cite{kim2018interpretability} & Enable interpretable visual or semantic concepts linked to model predictions & Limited by interpretability of visual or semantic features; do not address privacy concerns effectively\\

\hline
\end{tabular}
\label{tab:xai_summary}
\end{table}

\textcolor{black}{Another approach, Rényi Differential Privacy (RDP), extends the DP framework by quantifying privacy loss using Rényi divergence, which measures the statistical distance between probability distributions. RDP introduces a more flexible way to compose privacy guarantees over multiple queries by leveraging the additivity of Rényi divergence across independent mechanisms. This allows for tighter privacy bounds and improved analysis of iterative algorithms like SGD. Compared to traditional DP accounting, RDP can better balance noise addition and utility, making it a valuable tool for scenarios requiring iterative optimization. However, while RDP improves upon the flexibility of privacy composition, its interpretation requires a deeper understanding of Rényi divergence and its relationship to privacy budgets, potentially complicating its application in some domains~\cite{mironov2017renyi}.}

\subsection{\textcolor{black}{Explainable AI}}

\textcolor{black}{Integrating DP into AM processes enables manufacturers to secure sensitive data, supporting privacy-centric ML applications for monitoring and analysis. However, introducing noise into ML models, especially those operating as black-box systems, complicates the interpretation of how noise impacts model accuracy. Traditional XAI approaches, reviewed extensively in~\cite{akkem2023smart}, primarily focus on improving interpretability in various fields but often lack the mechanisms to handle DP-specific challenges, such as the predictable impact of DP noise on model accuracy. This limitation highlights a gap in existing XAI methods, particularly under stringent privacy constraints, thereby underscoring the need for more specialized frameworks like DP-HD.}

\textcolor{black}{To address this gap, the DP-HD framework we propose employs an innovative approach centered around an SNR metric. This metric quantifies the contribution of training data relative to DP noise, establishing a structured foundation for balancing privacy and accuracy. Inspired by advanced XAI techniques found in recent applications like crop recommendation systems~\cite{akkem2024streamlit}, DP-HD integrates an SNR-based interpretation layer to enhance transparency. This approach allows the framework to both monitor the effect of DP noise on model predictions and reliably estimate accuracy across privacy levels. By offering this quantifiable insight, DP-HD extends beyond conventional XAI limitations, providing a robust mathematical solution that clarifies the impact of DP noise in high-dimensional AM models while ensuring data privacy.}

\textcolor{black}{Within the broader field of XAI, a variety of techniques have been developed to make ML models more interpretable. Attribution methods, widely used for their adaptability and clarity, assess the contribution of input features to model predictions, particularly through post-hoc analysis methods~\cite{fan2021interpretability}. Saliency methods~\cite{simonyan2013deep} were among the first to apply backpropagation for explaining model predictions, with extensions like InputXGradient~\cite{shrikumar2016not} and others (GuidedBackpropagation~\cite{springenberg2014striving} and IntegratedGradients~\cite{sundararajan2017axiomatic}) further refining this approach. These gradient-based techniques provide insight by accessing model internals, a feature unavailable in black-box contexts.}

\textcolor{black}{In contrast, perturbation-based methods like Occlusion~\cite{zeiler2014visualizing} are model-agnostic, meaning they do not require access to specific model architectures. Such methods work by systematically altering inputs to observe the effect on model predictions. Other approaches, like Local Interpretable Model-Agnostic Explanations (LIME)~\cite{ribeiro2016should}, generate local surrogate models to explain predictions, while Shapley Additive Explanations (SHAP)~\cite{lundberg2017unified} apply game theory principles to offer mathematically grounded insights into feature importance.}

\textcolor{black}{Beyond attribution-based methods, other XAI techniques, including prototype-based~\cite{li2018deep}, patch-based~\cite{chen2019looks}, and concept-based methods~\cite{kim2018interpretability}, offer additional ways to interpret complex models. However, while these traditional XAI methods provide valuable interpretability, they often fall short of addressing DP noise effects on model accuracy, a critical aspect of balancing privacy and utility in high-stakes applications like AM.}

\textcolor{black}{The DP-HD framework offers a structured approach to balancing privacy and accuracy by introducing an SNR-based metric, making it well-suited for sensitive applications that require strict data privacy protections. While developed within the context of AM, the principles behind DP-HD can also be extended to other privacy-sensitive environments, such as healthcare, financial services, and government data management, where secure handling of confidential information is paramount. In these domains, where sensitive data like patient records, financial transactions, and personal identifiers must be protected, DP-HD’s ability to quantify and predict noise impact on model accuracy provides an advanced solution. This generalizability underscores the DP-HD framework's potential as a versatile tool for privacy-preserving ML across various sectors, ensuring robust protection without sacrificing model performance. Table~\ref{tab:xai_summary} summarizes the discussed XAI methods, highlighting their respective advantages and limitations.}

\section{Research Methodology}
\label{sec:Research Methodology}

\subsection{Hyperdimensional Computing}

\begin{figure}
    \centering
    \includegraphics[width=\textwidth]{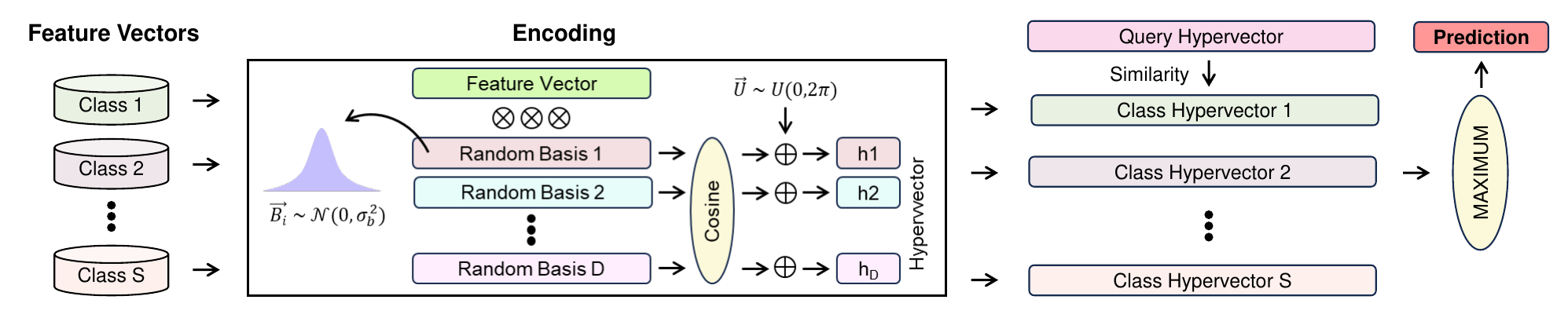}
        \caption{\textcolor{black}{Hyperdimensional computing framework, including encoding, training, inference, and retraining. In encoding, feature vectors are mapped to hypervectors, where each element, $h_i$, is generated by projecting a feature vector onto a random basis vector $\vec{B}_i$, sampled from a normal distribution with mean zero and variance $\sigma_b^2$. A uniform distribution, $\vec{U}$, is added to enhance the projection. Training aggregates hypervectors from the same class to form class hypervectors, inference compares query hypervectors for similarity, and retraining adjusts hypervectors based on misclassifications.}}
    \label{hd}
\end{figure}

Hyperdimensional Computing (HD) draws significant inspiration from the architecture of the human brain, utilizing high-dimensional vectors to simulate cognitive functions~\cite{piran2024privacy,hoang2024hierarchical,rescsanski2023anomaly}. This advanced approach allows HD to emulate brain-like methodologies for processing, analyzing, and storing information across diverse cognitive tasks. The HD framework, depicted in Figure~\ref{hd}, is structured into four key phases: encoding, training, inference, and retraining. Each phase is meticulously designed to model different aspects of cognitive processing.

The process begins with the encoding phase, where feature vectors are transformed into high-dimensional hypervectors, forming the foundation for the subsequent training phase. During training, hypervectors are aggregated to create distinct class representations, crucial for the model's ability to identify and classify new data during the inference phase. The cycle concludes with the retraining phase, which allows for continuous refinement and improvement of the model based on new data and insights. This dynamic process mirrors the brain's capacity for learning and adaptation, showcasing HD's ability to handle complex learning tasks with remarkable accuracy and flexibility.

\subsubsection{Encoding}
\label{sec:encoding}

\textcolor{black}{The encoding phase transforms feature vectors into high-dimensional hypervectors, ensuring a balanced data representation by distributing information uniformly across each element. As depicted in Figure~\ref{hd}, a feature vector \(\vec{F} = \{f_{j}\}_{j=1}^{J}\) is mapped into a hypervector \(\vec{H}\) by applying a dot product with a unique basis vector for each dimension. Enhanced with a uniformly distributed random sequence \(\Vec{U} = \{u_{i}\}_{i=1}^{D}\) sampled from \(\mathit{U}(0, 2\pi)\), the result is given by \(h_{i} = \cos(\Vec{F} \cdot \Vec{B}_{i} + u_i)\), where \(\Vec{B}_{i}\) represents Gaussian random vectors with mean zero and SD \({\sigma}_{b}\). These basis vectors remain consistent throughout, enabling continuous feature-to-hypervector conversion for tasks like anomaly detection~\cite{JalilPiran2023}.}

The choice of the SD denoted as \(\sigma_{\text{b}}\) within the Gaussian distribution used to generate the random basis vectors, \(\vec{B}_i\), is crucial in encoding. Altering the SD affects the encoding's nature, leading to either exclusive or inclusive strategies. An exclusive strategy, achieved with a larger \(\sigma_{\text{b}}\), broadens the dispersion of the basis vectors' values, enhancing the uniqueness of each hypervector and aiding in distinguishing them. This is advantageous in tasks requiring precise identification of unique patterns. Conversely, an inclusive strategy, characterized by a smaller \(\sigma_{\text{b}}\), results in a more concentrated distribution of basis vector values, making hypervectors more similar. This is beneficial for tasks needing generalization over similar features. Adjusting the SD parameter thus provides nuanced control over the encoding's specificity and generalizability, demonstrating HD's versatility in handling various classification challenges.

\subsubsection{Training}

The training phase begins with the creation of class hypervectors, denoted by $\vec{C}_s$ for each class $s$, with $s$ ranging from 1 to $S$ representing the total number of classes. Each class hypervector is formed by summing all related hypervectors for that class, as formulated: \begin{equation} 
\vec{C}_s = \sum \vec{H}^s
\label{eq:formClassHypervector}
\end{equation} 
where $\vec{H}^s$ represents the hypervectors associated with class $s$. This aggregation mechanism ensures that the features of the training samples from each class are combined.

By grouping hypervectors from the same class, the HD framework merges common characteristics and attributes within a high-dimensional space. This process ensures that the fundamental aspects of each class are encapsulated into a unified representation, which is essential for accurate and efficient classification. Leveraging the principle of superposition inherent to high-dimensional spaces, the HD framework enhances its capacity to distinguish between classes by analyzing their combined features.

Creating class hypervectors is a pivotal phase in the HD training process, laying the groundwork for the inference and retraining phases. This step enables HD to form a comprehensive understanding of each class's distinct high-dimensional profile, crucial for the model's pattern recognition and prediction capabilities.

\subsubsection{Inference}

During the inference phase, a hypervector's class association is determined by comparing the query hypervector to the class hypervectors established during training. Cosine similarity is used for this comparison, as it measures how closely two vectors align in direction within the high-dimensional space. The similarity is calculated as the dot product of two vectors, normalized by the product of their norms: 
\begin{equation}
\text{Cos}(\Vec{C}_{s}, \Vec{H}) = \frac{\Vec{C}_{s} \cdot \Vec{H}}{||\Vec{C}_{s}|| \cdot ||\Vec{H}||}
\label{eq:inference similarity}
\end{equation}

During inference, the HD model evaluates the query hypervector $\Vec{H}$ against each class hypervector using cosine similarity metrics. The class whose hypervector $\Vec{C}_{s}$ exhibits the highest similarity with $\Vec{H}$ is identified as the most probable class for $\Vec{H}$. This approach enables the HD model to classify new data by leveraging the learned patterns and utilizing the spatial properties unique to high-dimensional spaces for class distinction.

The use of cosine similarity in HD's inference methodology ensures high accuracy and computational efficiency, making it highly adaptable for various classification tasks. Notably, the norm of the query hypervector, $||\Vec{H}||$, remains constant across all classes, allowing it to be excluded from detailed calculations. Similarly, the magnitude of each class hypervector, $||\Vec{C}_{s}||$, stays unchanged during inference because these vectors are precomputed during training. Consequently, the inference process is reduced to computing dot products between the query and class hypervectors, which is a fast operation. This swift processing makes HD exceptionally suitable for in-situ sensing in AM, where rapid and accurate data classification is critical.

\subsubsection{Retraining}

Retraining significantly enhances the HD model's accuracy and adaptability. This phase involves comparing hypervectors of training samples with previously established class hypervectors from the initial training phase, identifying and correcting inaccuracies to improve overall performance. If a hypervector, $\Vec{H}^{s}$, is misclassified into an incorrect class, $s^{'}$, instead of its actual class, $s$, the HD model updates to correct this error. The class hypervectors are adjusted based on the magnitude of the misclassification, as detailed:
\begin{equation}
\begin{aligned}
\Vec{C}_s  = \Vec{C}_s + \Vec{H}^{s}, \
\Vec{C}_{s^{'}} = \Vec{C}_{s^{'}} - \Vec{H}^{s}
\end{aligned}
\end{equation}

This recalibration effectively shifts the class hypervectors, integrating the misclassified hypervector, $\Vec{H}^{s}$, into its correct class, $s$, and removing it from the incorrect one, $s^{'}$. These adjustments bring the correct class's hypervector closer to $\Vec{H}^{s}$ and move it further from the misclassified hypervector, reducing the likelihood of future misclassifications.

The retraining process is inherently iterative, allowing the HD model to continuously refine its classification accuracy by addressing errors. Each update incrementally enhances the model's ability to capture nuanced data variations, increasing its efficiency and reliability. This ongoing optimization is invaluable in environments subject to data distribution shifts or the emergence of new patterns.

Retraining transforms the HD model into an increasingly effective classification tool, capable of adapting to evolving data landscapes without restarting the training process from scratch. This adaptability, combined with HD's inherent efficiency, makes it an outstanding solution for real-time data processing and decision-making across varied applications.
\begin{table}[h!]
\centering
\color{black}
\caption{\textcolor{black}{Notation and description of variables used in the differential privacy-hyperdimensional computing framework.}}
\begin{tabular}{c c}
\hline
Notation & Description   \\
\hline

\(\vec{F} = \{f_{j}\}_{j=1}^{J}\) &  Feature vector  \\
$J$ & Feature dimension \\
\(\vec{H}\)  & Hypervector   \\
$D$ & Hypervector size  \\
$\vec{C}_s$ &  Class hypervector assigned to class $s$  \\
$\tilde{\vec{C}}_{s}$ & Noisy class hypervector  \\
$\vec{H}_q$ & Query hypervector  \\
$\{\vec{B}_{i}\}_{i=1}^{D}$ & Random basis vectors with a distribution of $\mathcal{N}(0,\sigma_{b}^2)$   \\
$\sigma^{*}_{b}$ & Optimal standard deviation of the random basis vectors\\
$\mathcal{N}(0, \Delta g^2 \sigma_{dp}^2)$&   Differential privacy noise \\
$\Delta g$ & Sensitivity of the model  \\
$g$ & AI model  \\
$\sigma_{dp}$ & Noise level   \\
$\epsilon$  & Privacy budget  \\
$\delta$ & Privacy loss threshold  \\
$M$ & Secure AI model  \\
$I_{1} and I_{2}$ & Two neighboring datasets differ in exactly one sample  \\

$N$ & The number of training samples \\
$T$ & The number of training epochs \\
$\mu_c$ & Average similarity of training hypervectors  \\
$\sigma_c$ & Standard deviation of the similarity among training hypervectors \\

$ \hat{\vec{F}}$ & Reconstructed feature vector \\
$\vec{\xi}$ & Encoder noise \\
$\Sigma$ & Variance of the decoder noise\\
 \hline
\end{tabular}
\label{table:notation}
\end{table}

\subsection{Differential Privacy}

In this study, we tackle the crucial issue of safeguarding confidentiality in in-situ sensing systems within AM. A primary focus is mitigating the risks posed by model inversion attacks. Traditional systems often unintentionally reveal sensitive information due to their extensive data accessibility requirements. To counteract this, we have implemented DP mechanisms. These mechanisms ensure that only modified outcomes, adjusted according to DP standards, are accessible instead of the original unmodified data. This approach significantly enhances security and reduces the risk of privacy violations from model inversion attacks, thereby protecting sensitive data and strengthening the integrity of AM models against security threats. The notation of the variables in this work is provided in Table~\ref{table:notation}.

DP bolsters the security framework of AM models by limiting access to raw data and permitting only the dissemination of results altered via DP techniques. DP operates by deliberately adding noise to obscure the original data, a process governed by two key metrics: the privacy budget ($\epsilon$) and the privacy loss threshold ($\delta$). These parameters are crucial in determining the intensity of the noise and the level of privacy protection, significantly reducing the likelihood of inferring individual data points. This method ensures that the privacy of data subjects is maintained without compromising the utility of the data for analysis, offering a robust solution to the challenge of protecting sensitive information in AM models from potential model inversion attacks.

\textbf{Definition 1.} To illustrate the concept of DP in the context of AM, consider two datasets, \(I_1\) and \(I_2\), which are identical except for one differing data point. A predictive model \(M\) satisfies \((\epsilon, \delta)\)-DP if, for any such pair of datasets \(I_1\) and \(I_2\), the following inequality holds:
\begin{equation}
\mathbb{P}[M(I_1)] \le e^{\epsilon} \cdot \mathbb{P}[M(I_2)] + \delta
\end{equation}
This criterion ensures that the presence or absence of a single data entry does not significantly influence the model's output, thereby protecting the confidentiality of individual data points.

\textbf{Definition 2.} In predictive modeling applications within AM, the Gaussian mechanism is a key technique for enforcing DP. This approach involves adding noise to the model's output to mask the influence of any single data point. Consider a function \(g: I \to \mathbb{R}^{D}\), which manipulates or transforms data within the model. The Gaussian mechanism, characterized by the noise level \(\sigma_{dp}\), is implemented as follows:
\begin{equation}
M(I) = g(I) + \mathcal{N}(0, \Delta g^2 \sigma_{dp}^2)
\label{eq:eq_dp}
\end{equation}

\textbf{Definition 3.} The term \(\Delta g\), known as the sensitivity of \(g\), represents the maximum expected difference in \(g\)'s output when comparing any two datasets, \(I_1\) and \(I_2\), that differ by only one element:
\begin{equation}
\Delta g = \max_{I_1, I_2} \|g(I_1) - g(I_2)\|
\end{equation}

\textbf{Theorem 1.} To achieve (\(\epsilon\), \(\delta\))-DP, noise level, \(\sigma_{dp}\), must satisfy the following condition:
\begin{equation}
\sigma_{dp} > \sqrt{2 \ln \frac{1.25}{\delta}} \cdot \frac{1}{\epsilon}
\label{eq:sigma_dp}
\end{equation}

\begin{algorithm}
\caption{Differential privacy-hyperdimensional computing framework.}
\label{alg:dphd}

\begin{algorithmic}[1]
\Statex \hspace{0em} \textbf{Input:}  $\{{\vec{F}}_{n}\}_{n=1}^{N}$, $\epsilon$, and $\delta$ \Comment{Training samples, privacy budget, and  privacy loss threshold}
\Statex \hspace{0em} \textbf{Output:}  $\{\tilde{{\vec{C}}}_{s}\}_{s=1}^{S}$ \Comment{Secured class hypervectors}

\Statex \hspace{0em} \textbf{Function Encoding:}
\State \hspace{0em} \textbf{for} $i \leq D$ \textbf{do}
\State \hspace{1.5em} $h_i \leftarrow \cos({\vec{F}} \cdot \Vec{B}_{i} + u_i)$ \Comment{Transform feature vectors into hypervectors}
\State \hspace{0em} \textbf{end for}
\State \hspace{0em} \textbf{Return} $\vec{H} \leftarrow [h_1,h_2,...,h_D]$

\Statex \hspace{0em}  \textbf{Function Training:}
\State \hspace{0em} \textbf{for} $s \in \{1, 2, \ldots, S\}$ \textbf{do}
\State \hspace{1.5em} ${\vec{C}}_{s} \leftarrow \sum \vec{H}^s$ \Comment{Aggregate hypervectors to form class hypervectors}
\State \hspace{0em} \textbf{end for}
\State \hspace{0em} \textbf{for} $epoch \leq T$ \textbf{do}
\State \hspace{1.5em} \textbf{if} Any misprediction 
\State \hspace{3em} $\Vec{C}_s  \leftarrow \Vec{C}_s + \Vec{H}^{s}$  \Comment{Correct  mispredictions}
\State \hspace{3em} $\Vec{C}_{s^{'}} \leftarrow \Vec{C}_{s^{'}} - \Vec{H}^{s}$
\State \hspace{1.5em} \textbf{end if}
\State \hspace{0em} \textbf{end for}
\State \hspace{0em} \textbf{Return} $\{{\vec{C}}_{s}\}_{s=1}^{S}$

\Statex \hspace{0em}  \textbf{Function Inference:}
\State \hspace{0em} \textbf{for} $q \leq Q$ \textbf{do}
\State \hspace{1.5em} $\text{Cos}(\Vec{C}_{s}, \Vec{H}_{q}) = \frac{\Vec{C}_{s} \cdot \Vec{H}_{q}}{||\Vec{C}_{s}|| \cdot ||\Vec{H}_{q}||}$
\State \hspace{1.5em} Assign class with highest similarity to $\Vec{H}_{q}$
\State \hspace{0em} \textbf{Return} Accuracy $\leftarrow \frac{\text{Correctly classified samples}}{Q}$

\Statex \hspace{0em}  \textbf{Function Privacy:}
\State \hspace{0em} $\sigma_{dp} \leftarrow  \sqrt{2 \ln \frac{1.25}{\delta}} \cdot \frac{1}{\epsilon}$
\State \hspace{0em} $  \tilde{\vec{C}}_{s} \leftarrow  \vec{C}_{s} + \mathcal{N}(0,\Delta g^2 \sigma_{dp}^2)$ \Comment{Add Gaussian noise to class hypervectors}  
\State \hspace{0em} \textbf{Return} $\{{\tilde{\vec{C}}}_{s}\}_{s=1}^{S} $

\Statex \hspace{0em} \textbf{Main DP-HD:}
\State \hspace{0em} $ \{{\vec{B}}_{i}\}_{i=1}^{D} \leftarrow  \mathcal{N}(0,\sigma_b^{2})$  \Comment{Generate random basis vectors}
\State \hspace{0em} $\{{\vec{H}}_{n}\}_{n=1}^{N} \leftarrow$ \textbf{Encoding(}$\{{\vec{F}}_{n}\}_{n=1}^{N}$,$\{{\vec{B}}_{i}\}_{i=1}^{D}${)}  
\State \hspace{0em} $\{{\vec{C}}_{s}\}_{s=1}^{S} \leftarrow$ \textbf{Training(}$\{{\vec{H}}_{n}\}_{n=1}^{N}${)}  
\State \hspace{0em} $\Delta g = \max \|\vec{H}_n\|$
\State \hspace{0em} $\{{\tilde{\vec{C}}}_{s}\}_{s=1}^{S}$ $\leftarrow$ \textbf{Privacy(}$\{{\vec{C}}_{s}\}_{s=1}^{S},\delta,\epsilon,\Delta g${)}  \Comment{Secured HD model}

\end{algorithmic}

\end{algorithm}

\subsection{\textcolor{black}{Differential Privacy-Hyperdimensional Computing}}
\label{sec:Differential Privacy-Hyperdimensional Computing}

\textcolor{black}{In this work, we present the DP-HD model, a novel approach in AM that integrates DP. This model excels in real-time monitoring and enables for precise mathematical adjustments of noise levels, crucial for safeguarding privacy without compromising accuracy, as illustrated in Algorithm~\ref{alg:dphd}. Ensuring the confidentiality of training data in AM is vital to prevent potential security breaches. To address this, Gaussian noise is strategically injected into the class hypervectors, thereby preventing the extraction of individual training samples by unauthorized entities.}

\textcolor{black}{During the training phase of the HD model, DP is implemented by making the class hypervectors differentially private through the injection of Gaussian noise, as depicted in Equation~\eqref{eq:noisy_chv}. This approach effectively reduces the risk of an adversary inferring specific details about the original feature vectors from the class hypervectors. The optimal \(\sigma_{dp}\) is determined using Equation~\eqref{eq:sigma_dp}, which aims to balance data privacy and the operational performance of our predictive model in the AM setting. In our specific case, we set \(\delta\) to \(10^{-4}\), based on the principle that \(\delta\) should be smaller than the inverse of the dataset size. The selection of the appropriate noise level, which is crucial for enhancing the model's security, depends on the DP-HD model's sensitivity. This sensitivity is determined by identifying the maximum norm across all encoded training samples, as shown in Equation~\eqref{eq:deltag}. A higher sensitivity necessitates the incorporation of greater noise levels, thus reinforcing the model's security. Therefore, the highest sensitivity observed across all training samples is used as a comprehensive sensitivity indicator, ensuring robust privacy measures across the dataset.}
\textcolor{black}{\begin{equation}
    \tilde{\vec{C}}_{s} =  \vec{C}_{s} + \mathcal{N}(0,\Delta g^2 \sigma_{dp}^2)
    \label{eq:noisy_chv}
\end{equation}
\vspace{-2em} 
\begin{equation}
    \Delta g = \max \|\vec{H_n}\|
    \label{eq:deltag}
\end{equation}}
\textcolor{black}{Another crucial parameter that influences the privacy level and sets the noise level is the privacy budget. The privacy budget can be selected to predict the impact of noise on the model's performance. In Section~\ref{sec:explainability}, we delve into the explainability of the DP-HD mechanism. We begin by assessing the effect of noise and training samples on the model's performance through an examination of its memorization capability. We then define the SNR, which is derived from the ratio of the signal (training samples) contribution to the model's performance against the noise contribution. This metric enables us to forecast how noise levels affect accuracy, allowing us to set specific privacy and performance levels concurrently. Following this, we explore the decodability of DP-HD. We analyze an analytical method for extracting information from a hypervector by subtracting two secure class hypervectors generated from datasets differing by only one sample. Moreover, in Section~\ref{sec:Encoding Strategy}, we outline an algorithm to find the optimal encoding strategy to balance privacy and performance. The encoding phase is critical to the model's success, as it determines how data is mapped from feature space to hyperspace.}

\subsubsection{\textcolor{black}{Explainability}}
\label{sec:explainability}

\textcolor{black}{Memorization in DP-HD occurs through the bundling operation during the training phase. As described earlier, the memorized vector is given by Equation~\eqref{eq:formClassHypervector}. To ensure DP during training, Gaussian noise with an SD of $\Delta g\sigma_{dp}$ is added to each component of the class hypervectors, as shown in Equation~\eqref{eq:noisy_chv}. This noise plays a crucial role during the inference process, where the algorithm attempts to recall one of the memorized hypervectors. In the presence of noise, the inference relies primarily on the similarity inner product between the secured class hypervector $\tilde{\vec{C}}_{s}$ and the query hypervector $\vec{H_q}$:
\begin{align}
\tilde{\vec{C}}_{s} \cdot \vec{H_q} = \vec{C}_s \cdot \vec{H_q} + \mathcal{N}(0, \Delta g^2 \sigma_{dp}^2) \cdot \vec{H_q}
\end{align}}

\textcolor{black}{Our goal is to analyze this noise and understand its effect on accuracy. For a better qualitative understanding, we work with a quantized hypervector such that $\vec{H_q}$ has components $\pm 1$. Consequently, the sensitivity is given by $\Delta g = \sqrt{D}$, where $D$ is the hypervector dimension. The similarity of the noise term can then be written as:
\begin{align}
\mathcal{N}(\vec{0}, \vec{1} \Delta g^2 \sigma_{dp}^2) \cdot \vec{H_q} \approx \mathcal{N}(0, D \Delta g^2 \sigma_{dp}^2) = \mathcal{N}(0, D^2 \sigma_{dp}^2)
\end{align}}

\textcolor{black}{Thus, the SD of the noise term is given by $\sigma_{noise} = D \sigma_{dp}$. Next, we examine the term contributed by the signal similarity $\vec{C}_s \cdot \vec{H_q}$. We assume that the average similarity between the same class elements follows an approximate Gaussian distribution with mean $\mu_c$ and SD $\sigma_c$. Therefore, the signal similarity can be expressed as:
\begin{align}
\vec{C}_s \cdot \vec{H_q} = \sum \vec{H}^s \cdot \vec{H_q} \approx \mathcal{N}(N \mu_c, N \sigma_c^2)
\end{align}}

\textcolor{black}{The signal distribution is Gaussian with mean $\mu_s = N \mu_c$ and SD $\sigma_s = \sqrt{N} \sigma_c$. For accurate classification, the signal distribution must be well separated from the noise distribution. The SNR can be defined as:
\begin{align}
{\rm SNR} = \frac{N \mu_c}{D \sigma_{dp}} = \frac{N \mu_c \epsilon}{D \sqrt{2 \ln \frac{1.25}{\delta}}}
\label{eq:snr}
\end{align}}

\textcolor{black}{This relationship allows us to quantify the impact of noise on the system's accuracy. A higher SNR indicates better accuracy by emphasizing the signal's dominance over noise during the inference stage. This equation also shows that increasing $\epsilon$ boosts the SNR, resulting in higher accuracy. However, it also reduces the noise level, thereby decreasing the privacy level according to Equations~\eqref{eq:sigma_dp} and \eqref{eq:noisy_chv}. Therefore, we select an $\epsilon$ value that balances the required privacy and model performance levels.}

\textcolor{black}{Moreover, as we can see, with a very large dimension $D$, the SNR ratio significantly decreases. The only way to counter this effect is to increase $\sigma_{dp}$, which is effectively done through a loss of DP by increasing $\epsilon$ and $\delta$. However, an alternate method of increasing the SNR is by simply increasing the dataset size $N$. Thus, having more training data reduces the effect of DP on the quality of the model. This can be intuitively understood since the data is sampled from the same underlying data distribution, the more dense the training sample is, the less the effect of removing a single training point becomes. In the limit of small ${\rm SNR}$, for good performance the width of the signal distribution should be much smaller than the width of the noise distribution.}

\textcolor{black}{Additionally, we analyze the effect of retraining the model. During retraining, noise must be added at each pass to ensure updates remain differentially private. The noise scales as $\sqrt{T}$, where $T$ is the number of retraining passes, while the signal scales linearly with $T$. Thus, the SNR during retraining is given by:
\begin{align}
    {\rm SNR}_{\rm retrain} = \frac{\sqrt{T}N\mu_c\epsilon}{D\sqrt{2 \ln \frac{1.25}{\delta}}}
\end{align}
This relationship reduces the impact of noise on the inference process and improves model quality.}

\textcolor{black}{Decodability is the next concept we discuss. We consider a scenario with two noisy class hypervectors that differ by only one sample, $\vec{H}$. The difference between these class hypervectors can be expressed as:
\begin{align}
    \tilde{\vec{C}}_{1} - \tilde{\vec{C}}_{2} = \vec{H} + \mathcal{N}(0,\Delta g^2 \sigma_{dp}^2)
\end{align}}

\textcolor{black}{The sample $\vec{H}$ encodes a feature vector, $\vec{F}$, using the equation $h_{i} = \cos(\Vec{F} \cdot \Vec{B}_{i})$. To retrieve the feature vector, we assume that $\vec{F}\cdot \vec{B}_i$ lies within the interval $[0, 2\pi)$. Our objective is to solve the set of equations $\vec{F}\cdot \vec{B}_i = v_i$, where $v_i=\cos^{-1}(h_i)$, to determine $\vec{F}$. This problem can be framed as a matrix equation. By grouping the $D$ components into sets of size $J$, where $J$ is the feature dimension, we can represent the equation for the feature vector as:}
\textcolor{black}{
\begin{align}
    \vec{v}_i = A_i\vec{F} + \vec{\xi}_i
\end{align}}

\textcolor{black}{In this equation, $i$ labels the different groups, $\vec{v}_i$ is the vector of size $J$ containing the $v_i$ values, and $A_i$ is a matrix whose rows correspond to the vectors $\vec{B}_i$. The term $\vec{\xi}_i$ represents the noise vector resulting from the DP-preserving noise added to the encoder.}

\textcolor{black}{We can then construct an estimator for the feature vector as $\hat{\vec{F}}_i = (A_i)^{-1}\vec{v}_i$ for each $i$, and then average out the noise by simple calculating the mean value extracted from all the groups,
\begin{align}
    \hat{\vec{F}}=\frac{\sum_i\hat{\vec{F}}_i}{K}
\end{align}
where $K$ refers to the number of groups. By writing out $\vec{f}_i$ in terms of the feature vector and noise term, we have that the following relationship holds 
\begin{align}
    \hat{\vec{F}} = \vec{F} + \frac{\sum_i (A_i)^{-1}{\vec{\xi}_i}}{K}
\end{align}
The quality of the inversion process directly depends on the strength of the noise term in the above equation. The noise can be significantly scaled up if the matrix $A_i$ is non-invertible. The covariance of a single term in the noise sum is given by 
\begin{align}
    \Sigma^0 &= \mathbb{E}\left((A_i)^{-1}\vec{\xi}_i\left((A_i)^{-1}\vec{\xi}_i\right)^\dagger\right)  \\
    &= \mathbb{E}\left((A_i)^{-1}\vec{\xi}_i\vec{\xi}_i^\dagger(A_i^\dagger)^{-1} \right) \notag \\
    &=\sqrt{2}\mathbb{E}\left((A_i)^{-1}\vec{\xi}_i\vec{\xi}_i^\dagger(A_i^\dagger)^{-1} \right) \notag
\end{align}
and under the assumption that $\vec{\xi}_i$ is a Gaussian with mean $0$ and covariance $\Delta g \sigma_{dp}$, we have that}

\textcolor{black}{\begin{align}
    \Sigma^0 = {\sqrt{2}\Delta g \sigma_{dp}}\mathbb{E}\left((A_i^\dagger A_i)^{-1} \right) \approx {\sqrt{2}\Delta g \sigma_{dp}}
\end{align}}

\textcolor{black}{where in the last inequality we approximate $\mathbb{E}\left((A_i^\dagger A_i)^{-1} \right) \approx \left(\mathbb{E}\left(A_i^\dagger A_i\right)\right)^{-1}  =\mathbb{I}$, since the components of $A_i$ are independently sampled from the normal distribution. Thus, the estimate for the feature vector can be written as 
\begin{align}
    \hat{\vec{F}} = \vec{F}+ \mathcal{N}(\vec{0}, \Sigma)
\end{align}}

\textcolor{black}{where $\Sigma= \Sigma^0/K ={\sqrt{2}\Delta g \sigma_{dp}}/K $. Thus, as expected, the noise during the decoding process also depends linearly on the sensitivity of the encoder and the amount of differential private noise added to the encoder. For a better DP, we need to add more noise which will increase the error in the decoder. This error can be reduced by increasing the value of the sampled in $K$. However, note that since $\Delta g$ is proportional to $D$, and $K$ is proportional to $D$, the ratio $\Delta g/K$ is independent of $D$. Thus, increasing the dimension should not change the error in the decoding process.}

\begin{algorithm}
\caption{Finding the best encoding strategy.}
\label{alg:encoding}

\begin{algorithmic}[1]
\Statex \hspace{0em} \textbf{Input:} $\{{\vec{F}}_{n}\}_{n=1}^{N}$ \Comment{Training samples}
\Statex \hspace{0em} \textbf{Output:} $\sigma^{*}_b$ \Comment{The optimal standard deviation for the random basis}

\Statex \hspace{0em} \textbf{Main Find $\sigma^{*}_b$:}
\State \hspace{0em} $\{{\vec{F}}_{m}\}_{m=1}^{M}$, $\{{\vec{F}}_{q}\}_{q=1}^{Q}$ $\leftarrow \{{\vec{F}}_{n}\}_{n=1}^{N}$ \Comment{Select subsets of training samples for training and validation}

\State \hspace{0em} \textbf{for} $\sigma_b \in [\sigma_b^{\text{min}},\sigma_b^{\text{max}}]$ \textbf{do}
\State \hspace{1.5em} $\{{\vec{B}}_{i}\}_{i=1}^{D} \leftarrow \mathcal{N}(0,\sigma_b^2)$ \Comment{Generate random basis vectors}

\State \hspace{1.5em} $\{{\vec{H}}_{m}\}_{m=1}^{M} \leftarrow$ \textbf{Encoding(}$\{{\vec{F}}_{m}\}_{m=1}^{M}$, $\{{\vec{B}}_{i}\}_{i=1}^{D}${)}  

\State \hspace{1.5em} $\{{\vec{C}}_{s}\}_{s=1}^{S} \leftarrow$ \textbf{Training(}$\{{\vec{H}}_{m}\}_{m=1}^{M}${)}  

\State \hspace{1.5em} $\{{\vec{H}}_{q}\}_{q=1}^{Q} \leftarrow$ \textbf{Encoding(}$\{{\vec{F}}_{q}\}_{q=1}^{Q}$, $\{{\vec{B}}_{i}\}_{i=1}^{D}${)}  

\State \hspace{1.5em} $\Delta g = \max \|\vec{H}_m\|$

\State \hspace{1.5em} \textbf{for} $\epsilon \in [\epsilon_{\text{min}},\epsilon_{\text{max}}]$ \textbf{do}

\State \hspace{3em} $\{{\tilde{\vec{C}}}_{s}\}_{s=1}^{S}$ $\leftarrow$ \textbf{Privacy(}$\{{\vec{C}}_{s}\}_{s=1}^{S}$, $\delta$, $\epsilon$, $\Delta g${)} \Comment{Secured class hypervectors}

\State \hspace{3em} Accuracy $\leftarrow$ \textbf{Inference(}$\{{\tilde{\vec{C}}}_{s}\}_{s=1}^{S}$, $\{{\vec{H}}_{q}\}_{q=1}^{Q}${)} 

\State \hspace{1.5em} \textbf{end for}

\State \hspace{1.5em} $\sigma^{*}_b \leftarrow$ The highest accuracy and the lowest $\epsilon$

\end{algorithmic}
\end{algorithm}
\subsubsection{Encoding Strategy}
\label{sec:Encoding Strategy}

The encoding strategy plays a vital role in balancing privacy and performance in the DP-HD model. To achieve this balance, it is essential to find the optimal SD, denoted as \({\sigma}^{*}_{\text{b}}\), which ensures both the highest accuracy and the highest privacy. The value of \({\sigma}^{*}_{\text{b}}\) is influenced by the distribution of training samples and the distance between the centers of different classes. It is essential to note that \({\sigma}^{*}_{\text{b}}\) is not dependent on the DP-HD model parameters, such as the size of the hypervectors.

 \begin{figure}
    \centering
    \includegraphics[width=0.8\textwidth]{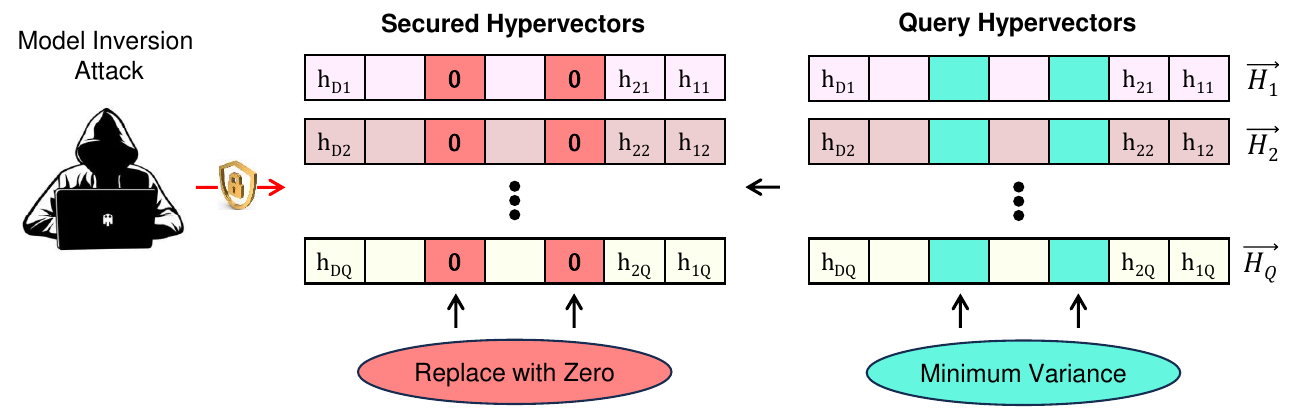}
        \caption{Securing inference phase: dropping low-variance dimensions.}
    \label{fig:infer_priv_drop_dim}
\end{figure} 

To determine \({\sigma}^{*}_{\text{b}}\), we select a subset of training samples that represent the distribution of the entire dataset for training, and another subset for validation. We then experiment with different \({\sigma}_{\text{b}}\) values to generate random basis vectors, as outlined in Algorithm~\ref{alg:encoding}. Using these values, we create secured class hypervectors with the training data for various privacy budget values. We then calculate the validation accuracy based on the model's performance in predicting the validation data labels. We store the accuracy for each pair of (\({\sigma}_{\text{b}}\),$\epsilon$). Finally, we select the SD for generating random basis vectors, \({\sigma}^{*}_{\text{b}}\), that corresponds to the highest accuracy and the lowest $\epsilon$.

\subsection{Inference Privacy}
\label{sec:Inference Privacy}

In Section~\ref{sec:Differential Privacy-Hyperdimensional Computing}, we discussed creating a secured model during the training phase. However, ensuring privacy during the inference phase is equally important. In the inference phase, a query hypervector is sent to the HD model, and a model inversion attacker could potentially decode the hypervector to uncover sensitive raw data. Therefore, it is crucial to introduce a secure mechanism for transmitting query hypervectors to the HD model.

In this section, we first define the concept of decodability of a hypervector, which attackers might exploit to infer the original feature vector. We then propose a novel method to secure hypervectors during the inference phase, thereby protecting sensitive information from potential attacks.

\subsubsection{Decodability of Hypervectors}
\label{sec:Decodability of Hypervectors}

Decodability is a challenging issue, mainly due to the use of the cosine activation function in HD encoders. Direct inversion of this function only reveals the value of $\vec{F} \cdot \vec{B}_i$ within $\mod 2\pi$. It is cryptographically infeasible to recover the feature vector $\vec{F}$ without access to a vast number of components, exponentially related to the feature dimension. Thus, if the distribution width of $\vec{B}_i$ is sufficiently large, feature recovery becomes impractical, ensuring the exclusivity of the data encoding process.

However, as discussed in Section~\ref{sec:Encoding Strategy}, the encoding strategy affects both privacy and model performance. The choice of an encoding strategy—whether exclusive or inclusive—is made to balance privacy and performance, rather than solely focusing on inference privacy. To provide some analytical insight, we assume that the inverse of the cosine function results in a value within the primary branch of $\cos^{-1}$. This scenario occurs when the values of both $\vec{B}_i$ and $\vec{F}$ are small, ensuring that $\vec{F} \cdot \vec{B}_i$ lies within the interval $[0, 2\pi)$. When $\vec{F} \cdot \vec{B}_i = v_i$ and $v_i = \cos^{-1} h_i$, the feature vector can be reconstructed from the random basis vectors using the following equation:

\begin{equation}
     f_{j} = \frac{\Vec{B}^{T}_{j} \cdot \Vec{V}}{D}
    \label{eq:inference decode}
\end{equation}

\begin{figure}[ht]
    \centering
    \begin{subfigure}[b]{0.33\textwidth}
        \centering
        \includegraphics[width=\textwidth]{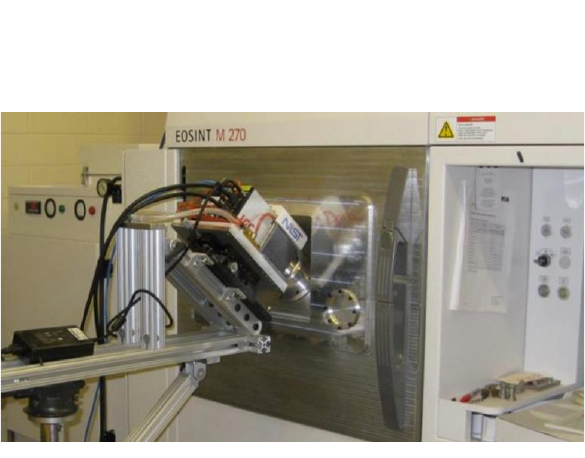}
        \caption{}
        \label{fig:datasetsub1}
    \end{subfigure}%
    \begin{subfigure}[b]{0.33\textwidth}
        \centering
        \includegraphics[width=\textwidth]{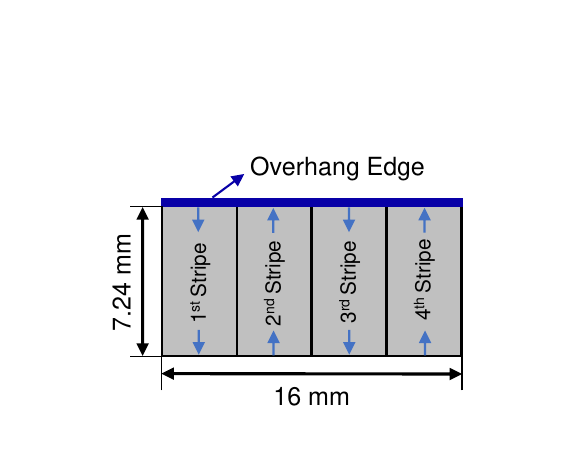}
        \caption{}
        \label{fig:datasetsub2}
    \end{subfigure}%
    \begin{subfigure}[b]{0.33\textwidth}
        \centering
        \includegraphics[width=\textwidth]{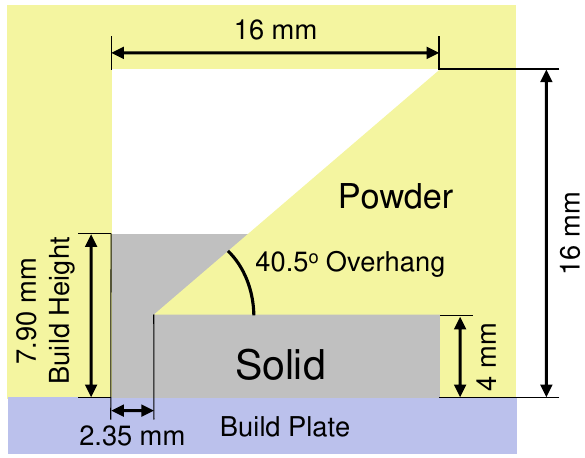}
        \caption{}
        \label{fig:datasetsub3}
    \end{subfigure}
        \caption{\textcolor{black}{The experimental setup, showing (a) the EOS M270 laser powder bed fusion additive manufacturing machine~\cite{lane2016thermographic}, (b) the top view schematic of the built part, and (c) the side view schematic with dimensions, including a 40.5-degree overhang.}}
    \label{fig:dataset}
\end{figure}

Even if $\vec{F} \cdot \vec{B}_i$ is not within the range $[0, 2\pi)$, attackers can use ML models to decode hypervectors and extract sensitive information. ML models typically perform a regression task to decode query hypervectors, where the hypervectors serve as inputs and the feature vectors are the outputs. The HD model is particularly vulnerable during the inference phase, as attackers might decode hypervectors to uncover confidential information in AM processes. Despite employing nonlinear encoding methods, ML models can still extract sensitive details from hypervectors. Therefore, implementing a privacy mechanism during the inference phase is crucial to protect this data. Thus, a secure mechanism is essential during the inference phase to prevent ML models from decoding hypervectors or accessing any sensitive information.

\subsubsection{Inference Privacy Mechanism}
\label{sec:Inference Privacy Mechanism}

In this section, we propose a secure mechanism for the inference phase to prevent ML models from decoding hypervectors. By reducing the dimension of hypervectors which is the size of ML models' input, we make it more difficult for attackers to infer sensitive information, thus enhancing privacy. To address privacy during the inference phase, we leverage the fact that not all dimensions of the query hypervectors contribute equally to the classification task. The primary operation during inference involves computing dot products between queries and class hypervectors, as described in Equation~\eqref{eq:inference similarity}. Some dimensions remain consistent across all query hypervectors, indicating that they only contain common information about the AM processes. Consequently, removing these dimensions will not significantly impact the HD model's performance.

Reducing the hypervector size complicates the task for attackers attempting to decode and extract feature information. In encoded queries, dimensions with low variance carry minimal meaningful information since they do not vary significantly across different classes. By setting these low-variance dimensions to zero, it becomes more challenging for attackers to infer information about the feature vectors, while the accuracy of the HD model remains largely unaffected. Based on Figure~\ref{fig:infer_priv_drop_dim}, we first identify the dimensions with low variance across all query hypervectors. Then, we replace these low-variance dimensions with zero. This approach ensures that the encoded queries remain private, protecting sensitive information during the inference phase.

\section{Experimental Design}
\label{sec:Experimental Design}

\textcolor{black}{Addressing privacy concerns in the dissemination and analysis of sensor data from AM processes is a paramount consideration. In industrial settings, protecting sensitive manufacturing data, such as part designs and proprietary process parameters, is crucial to maintaining competitive advantage and ensuring regulatory compliance. This study, therefore, not only focuses on the technical aspects of in-situ sensing and defect detection but also realizes the implementation of DP measures. These measures ensure that the valuable insights gained from high-speed camera data do not compromise the confidentiality of the manufacturing process or the design specifications of the parts being produced. This introduction sets the stage for a comprehensive investigation into the experimental designs and methodologies employed in this research. The subsequent sections delve into the specifics of sensor integration, data acquisition, and the analytical techniques used to achieve the dual objectives of enhancing manufacturing precision and ensuring data privacy in the realm of AM.}

\subsection{LPBF AM Machine Instrumentation and Setup}

\textcolor{black}{This section outlines the instrumentation of a commercial LPBF machine, specifically the EOS M270 model situated at the National Institute of Standards and Technology (NIST) as shown in Figure~\ref{fig:datasetsub1}. Integration involved outfitting the machine with a high-speed visible camera. The camera boasts a frame rate of 400 frames per second and a spectral response spanning 300-950 nm, facilitating real-time monitoring of the LPBF process. This monitoring entails capturing disparities in the thermal signature of the part as it undergoes layer-wise construction, enhancing quality assurance. Positioned within the build chamber's upper right corner, the high-speed visible camera is connected via data cables routed through a custom sealed through-port. Notably, precise measurements regarding the camera's orientation and distance were not conducted. The camera utilized features a Silicon-based array and boasts a resolution of 1.2 megapixels. Its placement inside the build chamber enables comprehensive imaging of the part's shape and surrounding spatter pattern. Images captured by the high-speed visible camera are windowed to dimensions of 256 pixels × 256 pixels and acquired at a rate of 1000 frames per second. These images serve as representative snapshots, depicting both overhang and bulk build features. The experimental setup replicates conditions in industrial LPBF processes, providing insights into defect detection and quality control practices~\cite{lane2016multiple}.}

\textcolor{black}{The example part was fabricated utilizing specific build parameters tailored for a commercial LPBF system. These parameters include a hatch distance of 0.1 mm, a stripe width of 4 mm, a stripe overlap of 0.1 mm, a powder layer thickness of 20 $\mu$m, a laser scan speed of 800 mm/s, and a laser power of 195 W during infill. Contour passes commence from distinct corners of the part, as indicated in Figure~\ref{fig:datasetsub2}, traversing counterclockwise along the part perimeter. Laser power settings for these passes vary, with pre-contour passes utilizing 100 W and post-contour passes employing 120 W, both at a scan speed of 800 mm/s. These parameters mirror those used in real-world industrial applications, emphasizing the practical relevance of the study’s findings. The study's focus lies on acquiring and initially processing sensor data, which entails exploring a single layer example from the 16 mm tall part. Data collection spans 45 different build layers, with the overhang delineated as the last two scan vectors prior to or just after edge formation, excluding pre- or post-contour scans. The remaining scans are attributed to the bulk volume of the part. A stripe pattern scan strategy is adopted (Figure~\ref{fig:datasetsub2} and \ref{fig:datasetsub3}), with laser scans over the overhang occurring four times per layer beyond 4 mm build height. The stripe orientation shifts by 90° between layers, and three example layers exhibit a vertical stripe pattern. The fabricated part and its features were chosen to reflect common challenges in industrial AM processes, such as overhang formation and thermal distortions. The test artifact, composed of nickel alloy 625, features a 40.5-degree overhang without additional support structures. Sensor data analysis focuses on three example build heights: 6.06 mm, 7.90 mm, and 9.70 mm, encompassing the formation of the overhang structure. Each layer consists of four 4 mm stripes alternated by 90°. The nickel alloy 625 powder originates from the machine vendor, with a mean powder diameter of 37.8 $\mu$m, utilized in conjunction with a custom alloy 625 substrate. A side-view schematic of the LPBF part (Figure~\ref{fig:datasetsub3}) illustrates the investigation into the effects of an overhanging structure. Data exploration in this paper centers on a single build layer at a height of 7.90 mm, although the overhanging structure extends to complete a 16 mm tall object.}

 \begin{figure}
    \centering
    \includegraphics[width=0.5\textwidth]{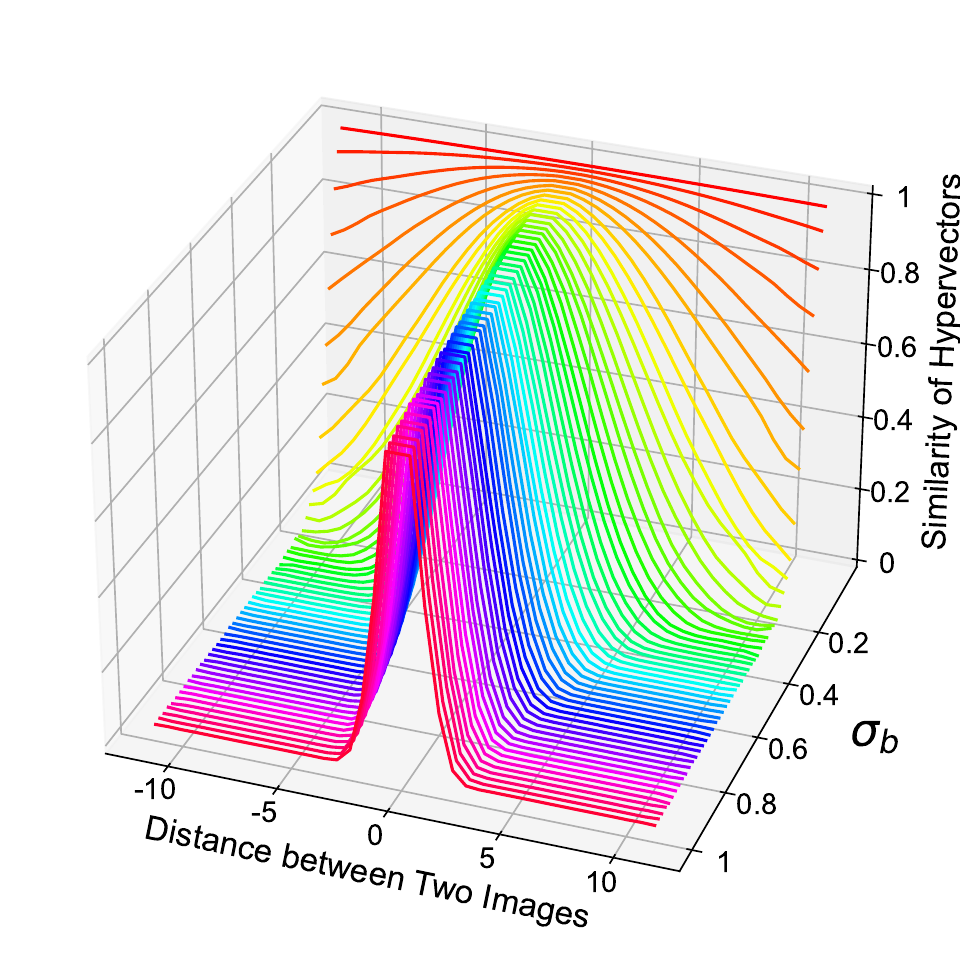}
    \caption{Comparative outcomes of exclusive and inclusive encoding.}
    \label{dis_sig_basis}
\end{figure}

\subsection{High-Speed Camera Data Collection}
\textcolor{black}{In the pursuit of advancing in-situ monitoring within the LPBF process, our experimental framework integrates a high-speed visible spectrum video camera, characterized by its rapid frame rate and high-resolution capabilities. Positioned strategically within the build chamber, the camera captures detailed thermal patterns as the laser fuses powder layers, creating a dynamic record of the melting process. This setup provides a realistic basis for investigating defect formation mechanisms and validating the effectiveness of DP-HD in industrially relevant scenarios. The specificity of the camera's placement and settings allows for an unprecedented view into the thermal dynamics at play, particularly in the formation of overhang features and the solidification of bulk areas.}

\textcolor{black}{The data acquisition phase is designed to ensure the capture of high-fidelity images, which are then windowed to a manageable size for processing. This optimization balances the need for detailed thermal data against the practical considerations of data storage and processing speed. The focus on high-speed camera data, among other sensor inputs, derives from its direct correlation with the thermal phenomena, offering a rich dataset for subsequent analysis.}

\subsection{Thermal Signature Classification}
\textcolor{black}{Building on the data captured, the study introduces a nuanced framework that delineates eight classes based on specific overhang and bulk characteristics across four distinct strips. This classification is not merely academic; it provides actionable insights into industrial AM challenges, such as identifying defects that could compromise part quality or structural integrity. Each class represents a unique thermal signature, corresponding to different combinations of overhang features and bulk regions, thereby enabling a detailed analysis of potential defects and irregularities.}

\subsection{Privacy-Enhanced Additive Manufacturing}
\textcolor{black}{As we navigate through the complexities of in-situ sensing and data analysis, the imperative of preserving privacy emerges as a critical concern. The integration of DP techniques in the processing of sensor data addresses this concern head-on, ensuring that while the data provides invaluable insights into the LPBF process, it does not expose sensitive information related to part designs or proprietary manufacturing techniques. This integration is particularly relevant for industries such as aerospace and automotive manufacturing, where data privacy is paramount. This dual focus on enhancing manufacturing precision while safeguarding data privacy forms the cornerstone of our approach, setting a new standard for research and practice in the field of AM.}

\begin{figure}
    \centering
    \begin{subfigure}[b]{0.5\textwidth}
        \centering
        \includegraphics[width=\textwidth]{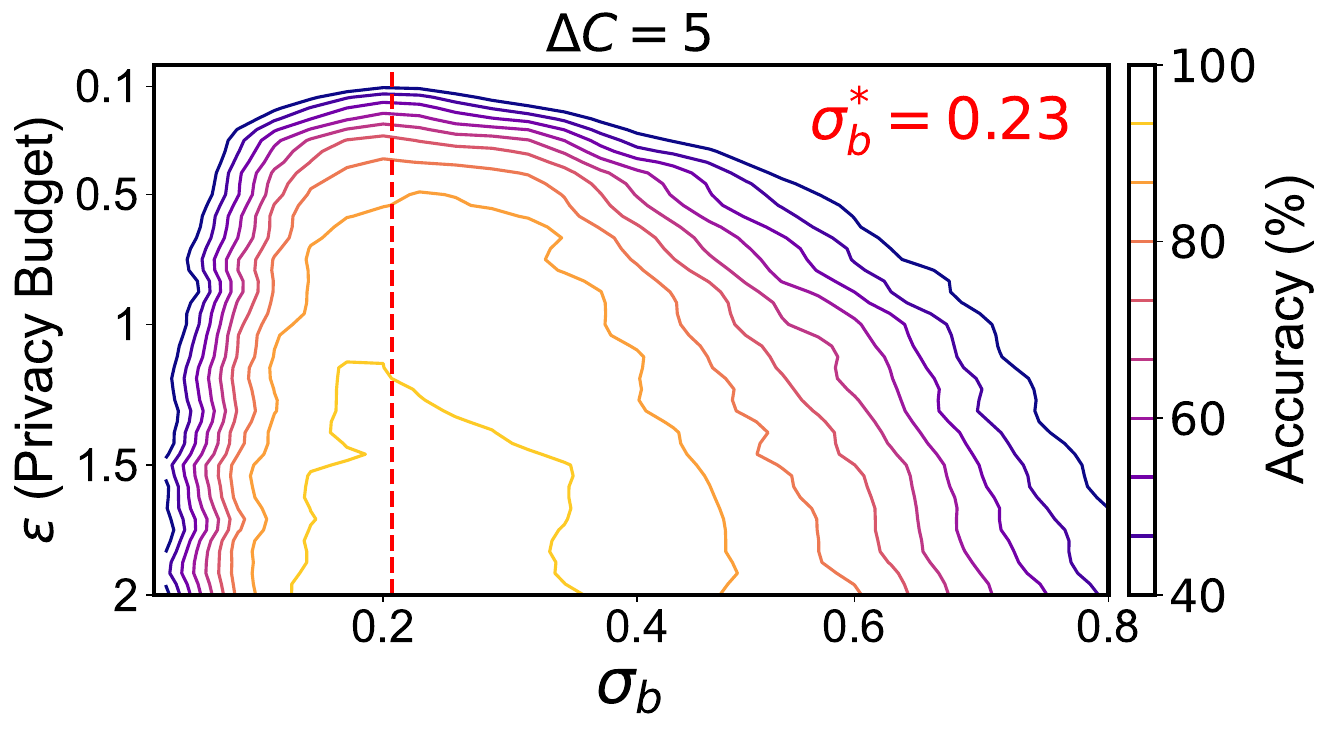}
        \caption{}
        \label{fig:acc_delta_c_5}
    \end{subfigure}%
    \begin{subfigure}[b]{0.5\textwidth}
        \centering
        \includegraphics[width=\textwidth]{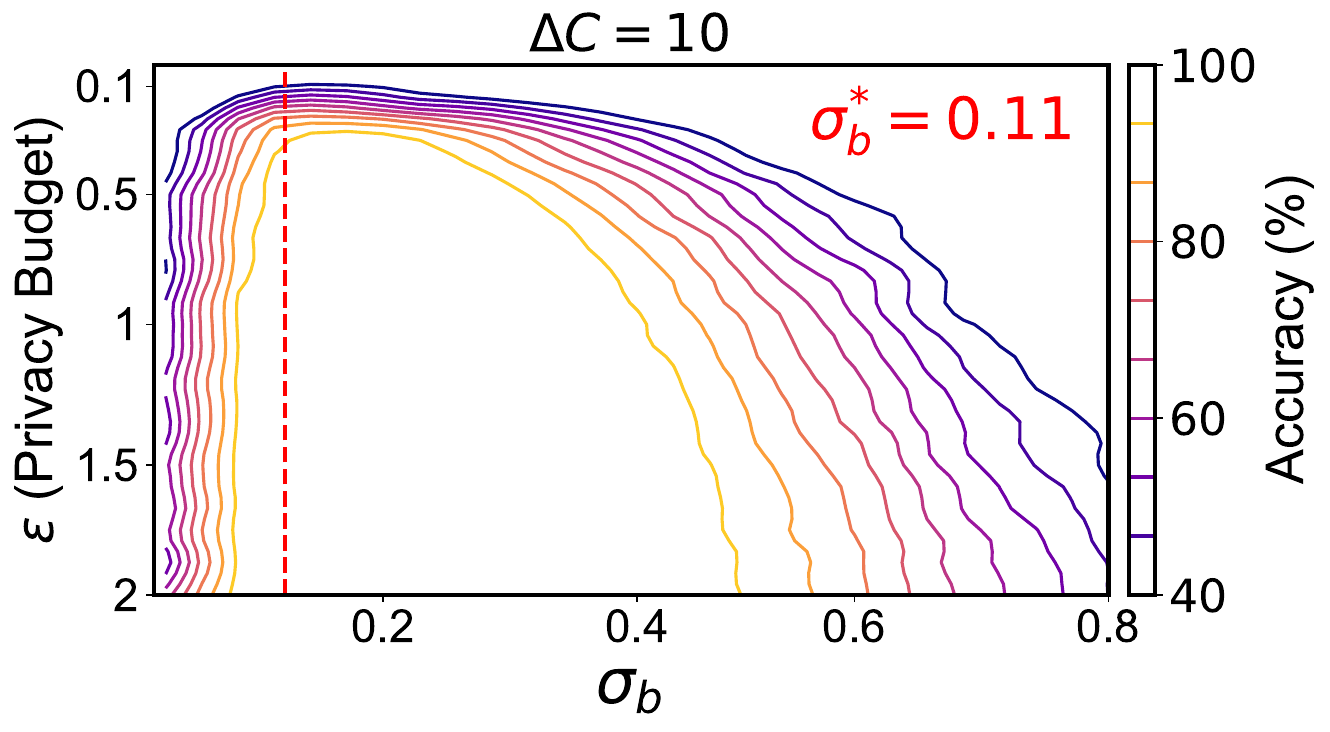}
        \caption{}
        \label{fig:acc_delta_c_10}
    \end{subfigure}
    \caption{Accuracy vs privacy balance when the distance between classes changed, (a) accuracy vs privacy balance with $\Delta C = 5$, (b) accuracy vs privacy balance with $\Delta C = 10$.}
    \label{fig:acc_delta_c}
\end{figure}

\section{Experimental Results} 
\label{sec:Experimental Results}

In this section, we thoroughly evaluate the DP-HD framework, focusing on both training and inference privacy. We begin by exploring encoding strategies, and examining the impact of exclusive and inclusive methods in mapping optical images to hyperspace. We then analyze how adding noise to class hypervectors affects the model's memorization capabilities and determine the optimal SD for generating random basis vectors. Additionally, we assess the explainability of the DP-HD model and benchmark its performance against the state-of-the-art DP-SGD and RDP mechanisms, implemented using Opacus, an open-source library for PyTorch that integrates DP into deep learning workflows~\cite{yousefpour2021opacus}. For inference privacy, we demonstrate the decoding of hypervectors to infer optical images from the LPBF process and present results on safeguarding query hypervectors. Finally, we compare the inference time of our proposed model with leading ML models in image classification tasks, highlighting the importance of efficient inference in in-situ sensing systems.

\begin{figure}[ht]
    \centering
    \begin{subfigure}[b]{0.45\textwidth}
        \centering
        \includegraphics[width=\textwidth]{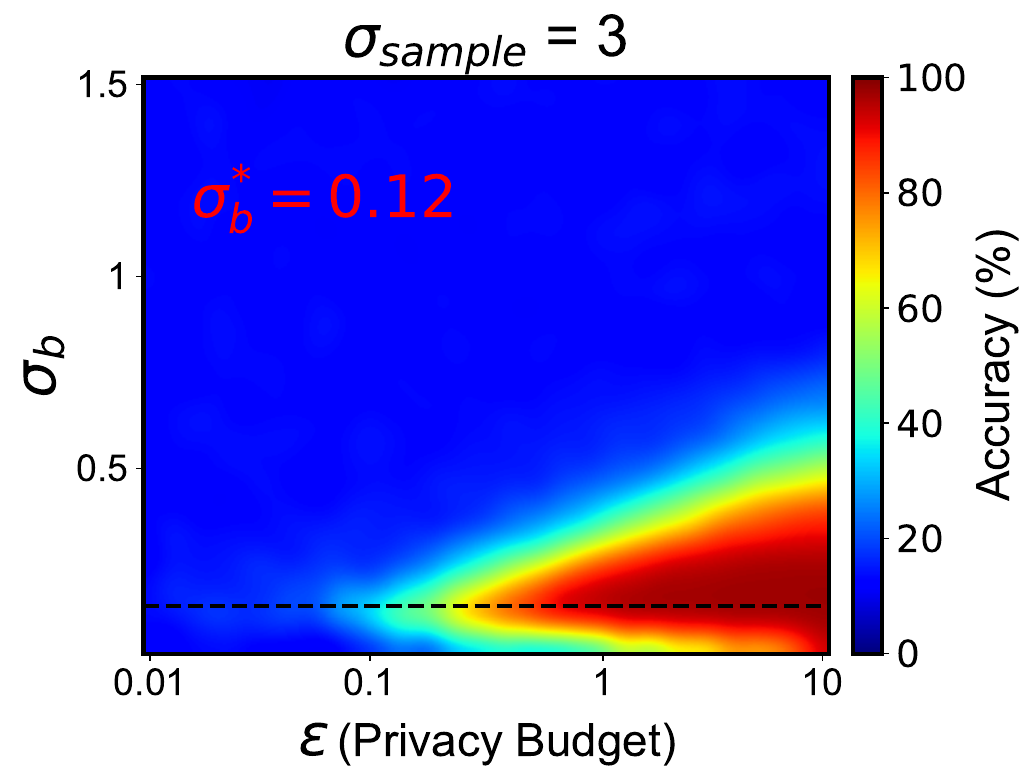}
        \caption{}
        \label{fig:sub1}
    \end{subfigure}%
    \begin{subfigure}[b]{0.45\textwidth}
        \centering
        \includegraphics[width=\textwidth]{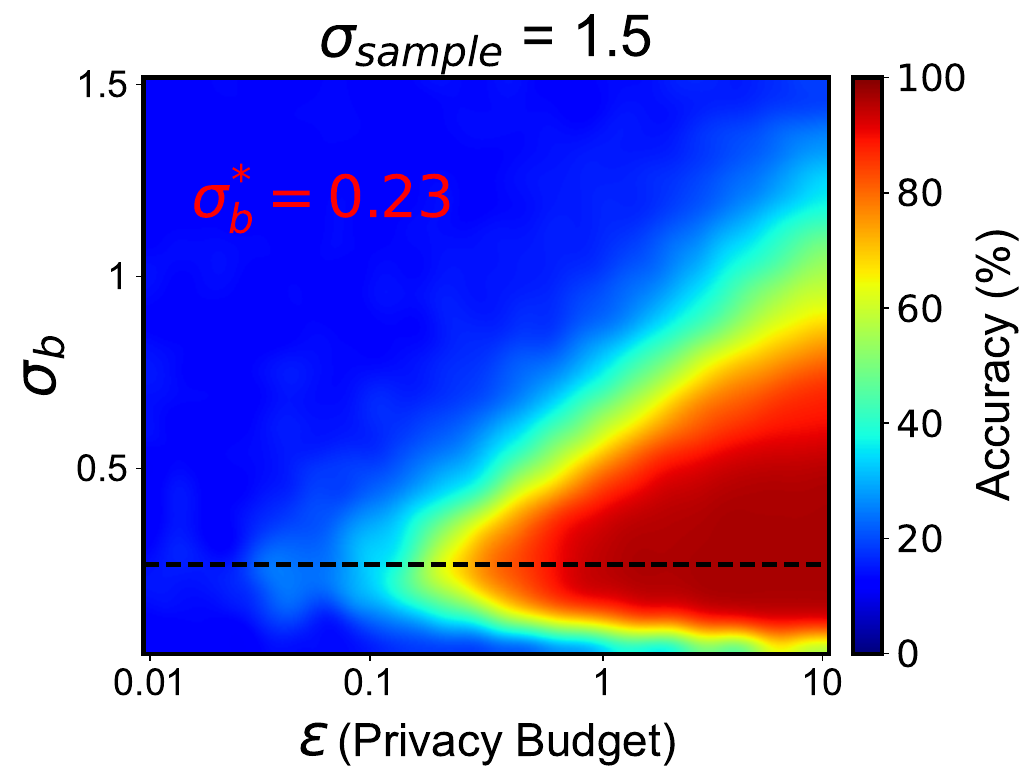}
        \caption{}
        \label{fig:sub2}
    \end{subfigure}\\ 
    \begin{subfigure}[b]{0.45\textwidth}
        \centering
        \includegraphics[width=\textwidth]{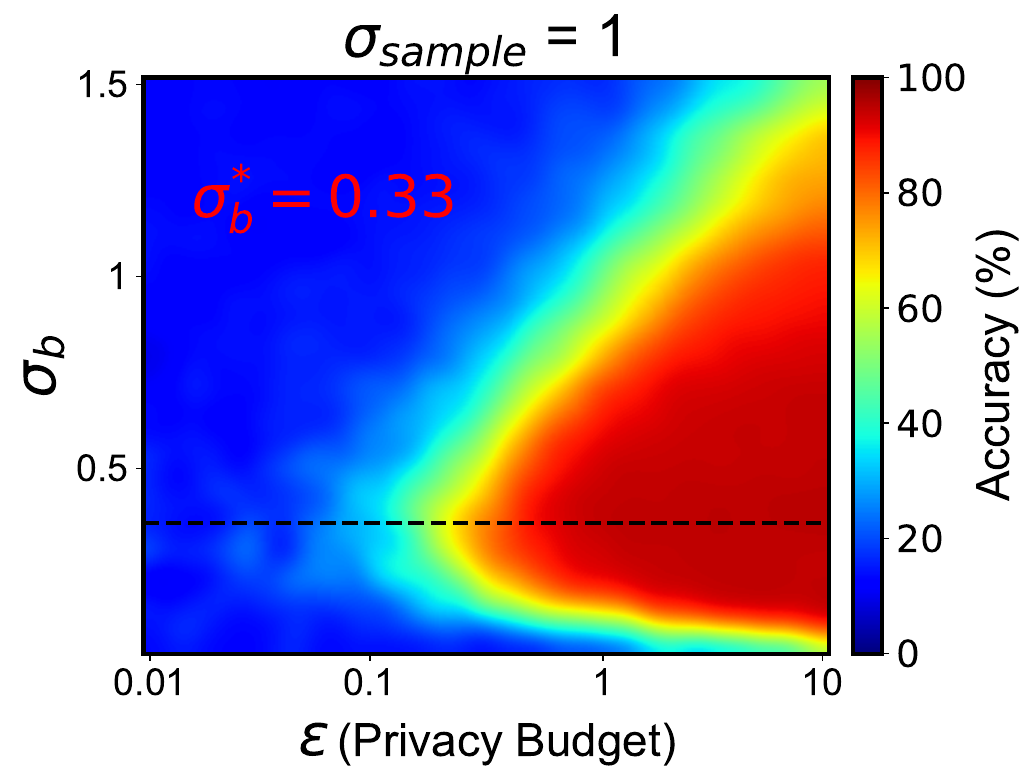}
        \caption{}
        \label{fig:sub3}
    \end{subfigure}%
    \begin{subfigure}[b]{0.45\textwidth}
        \centering
        \includegraphics[width=\textwidth]{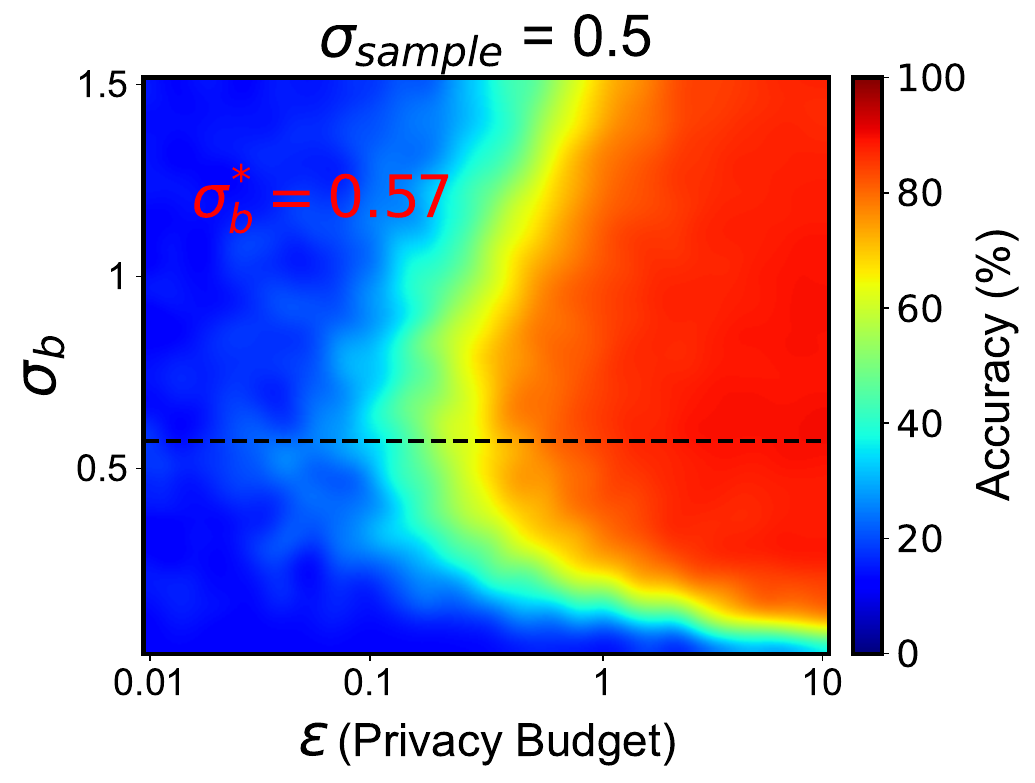}
        \caption{}
        \label{fig:sub4}
    \end{subfigure}
    \caption{Accuracy vs privacy balance for different sample spread in a feature space, (a) accuracy vs privacy balance with $\sigma_{sample} = 3$, (b) accuracy vs privacy balance with $\sigma_{sample} = 1.5$, (c) accuracy vs privacy balance with $\sigma_{sample} = 1$, and (d) accuracy vs privacy balance with $\sigma_{sample} = 0.5$.}
    \label{fig:acc_sig}
\end{figure}

\subsection{Training Privacy}

\subsubsection{\textcolor{black}{Sensitivity and Uncertainty Analysis}} 

\textcolor{black}{Identifying the optimal SD, denoted as \({\sigma}^{*}_{\text{b}}\), is essential for effective encoding processes. Figure~\ref{dis_sig_basis} shows how SD influences the effectiveness of encoding for optical images, illustrating the trade-off required between exclusive and inclusive encoding approaches. A smaller \({\sigma}_{\text{b}}\), for example, less than 0.1, leads to a significant level of similarity among hypervectors, even when the images are widely separated. This similarity often surpasses 0.5 under these conditions. On the other hand, as \({\sigma}_{\text{b}}\) increases, the similarity of hypervectors starts to correspond more closely with the actual physical distances between the images. At a \({\sigma}_{\text{b}}\) of 1, there is almost zero similarity in hypervectors for images that are significantly distanced. Such findings highlight the vital role of selecting a \({\sigma}_{\text{b}}\) that optimally balances the feature space compression and dilution, ensuring that the hypervectors are true representations of the original feature vectors.}

\textcolor{black}{In DP-HD, the optimal parameter, \({\sigma}^{*}_{\text{b}}\), achieves both higher accuracy and a more effective privacy budget, implying enhanced noise and improved privacy. For a synthetic dataset for a classification task, with each class's samples distributed in a Gaussian manner around class centers, characterized by an SD of \({\sigma}_{\text{sample}}\). These centroids maintain a uniform separation distance of \(\Delta C\). Figure~\ref{fig:acc_delta_c} illustrates how varying \(\Delta C\) affects \({\sigma}^{*}_{\text{b}}\). A \(\Delta C\) of 5, indicating proximity between classes, results in a \({\sigma}^{*}_{\text{b}}\) of 0.23. Increasing the separation to 10 changes \({\sigma}^{*}_{\text{b}}\) to 0.11, demonstrating the need for a more inclusive encoding strategy to generalize across similar features, as elaborated in Section~\ref{sec:encoding}. Additionally, the distribution spread of each class, \({\sigma}_{\text{sample}}\), also impacts \({\sigma}^{*}_{\text{b}}\). As shown in Figure~\ref{fig:acc_sig}, reducing \({\sigma}_{\text{sample}}\) results in an increase in \({\sigma}^{*}_{\text{b}}\), necessitating precise differentiation between encoded vectors through an exclusive encoding strategy explained in Section~\ref{sec:encoding}. \({\sigma}^{*}_{\text{b}}\) is 0.12, 0.23, 0.33, and 0.57 when \({\sigma}_{\text{sample}}\) is 3, 1.5, 1, and 0.5, respectively.}

\begin{figure}[ht]
    \centering
    \begin{subfigure}[b]{0.5\textwidth}
        \centering
        \includegraphics[width=\textwidth]{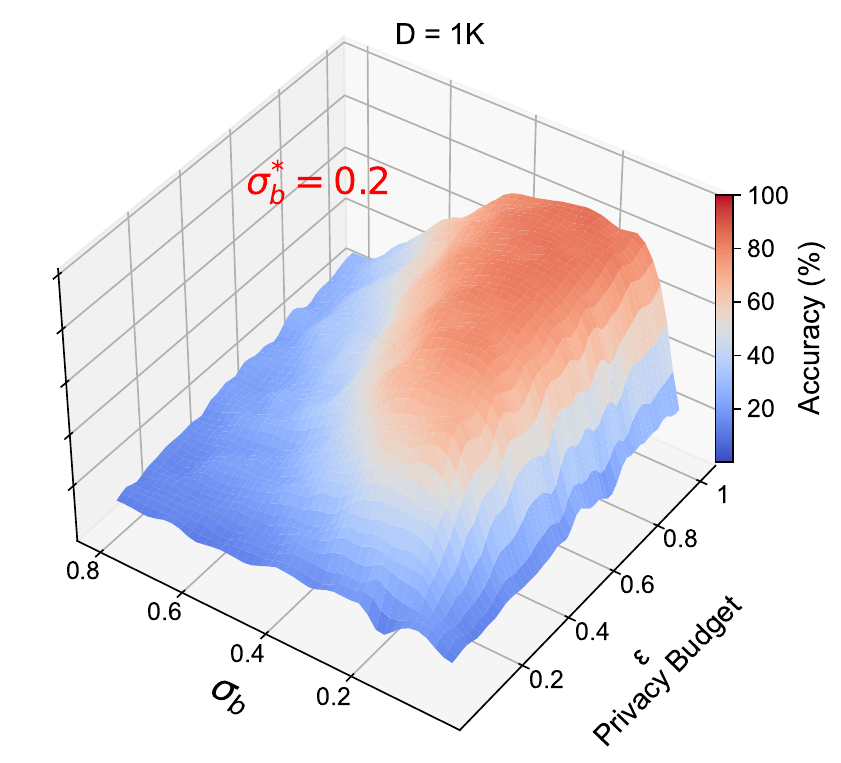}
        \caption{}
        \label{fig:sub1}
    \end{subfigure}%
    \begin{subfigure}[b]{0.5\textwidth}
        \centering
        \includegraphics[width=\textwidth]{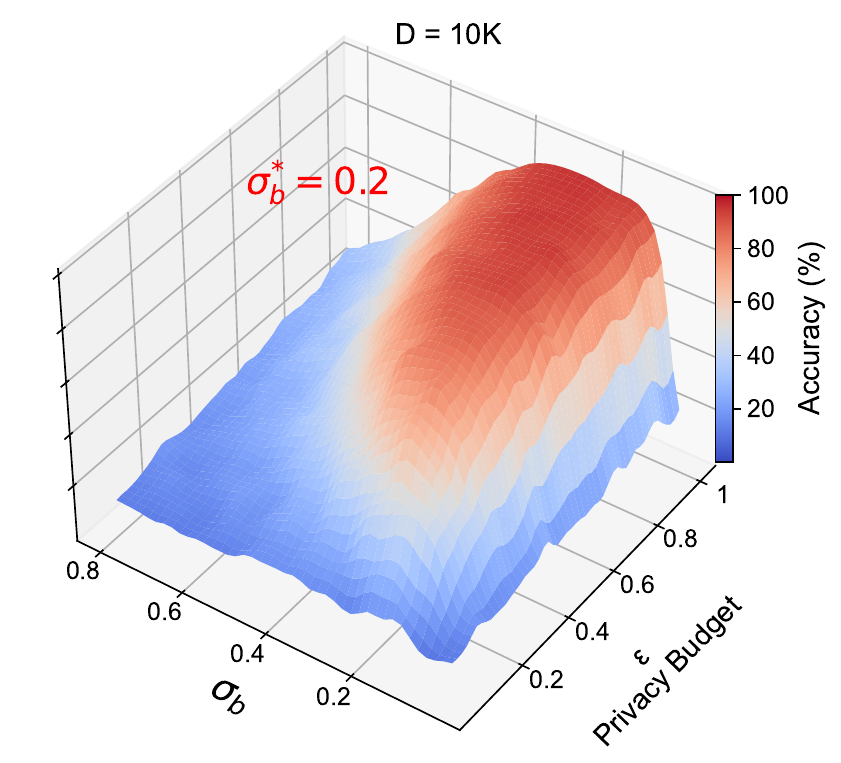}
        \caption{}
        \label{fig:sub2}
    \end{subfigure}\\ 
    \caption{Accuracy vs privacy balance for different hyperspace size, (a) accuracy vs privacy balance with $D = 1K$, and (b) accuracy vs privacy balance with $D = 10K$.}
    \label{fig:acc_d}
\end{figure}

\begin{table}[h!]
\centering
\caption{The effect of training size on the optimal encoding strategy and the privacy level when the accuracy is higher than 90\%.}
\begin{tabular}{c c c c}
\hline

Train Set Size & Accuracy & $\epsilon$ & $\sigma^{*}_{b}$    \\
\hline
1K &   & 0.15 & 0.22 \\
2K &   & 0.08& 0.22 \\
3K & >90\%  & 0.06 & 0.22\\
4K &   & 0.05 & 0.22\\
5K &    & 0.03& 0.22\\
\hline
\end{tabular}
\label{table:acc_train_size}
\end{table}

\textcolor{black}{Subsequently, we modified the size of the hypervector and the number of training samples, key factors in the DP-HD framework. Figure~\ref{fig:acc_d} demonstrates the impact on accuracy across various hypervector sizes. It becomes evident that \({\sigma}^{*}_{\text{b}}\) remains unaffected by variations in hypervector size. Likewise, as depicted in Table~\ref{table:acc_train_size}, the number of training samples also does not influence \({\sigma}^{*}_{\text{b}}\). Thus, we deduce that \({\sigma}^{*}_{\text{b}}\) is unrelated of these DP-HD parameters and the scale of training data. It is solely contingent upon the distribution characteristics of the feature vectors, such as the proximity between class centers and the spread of samples around these centers.}

\begin{figure}
    \centering
    \includegraphics[width=\textwidth]{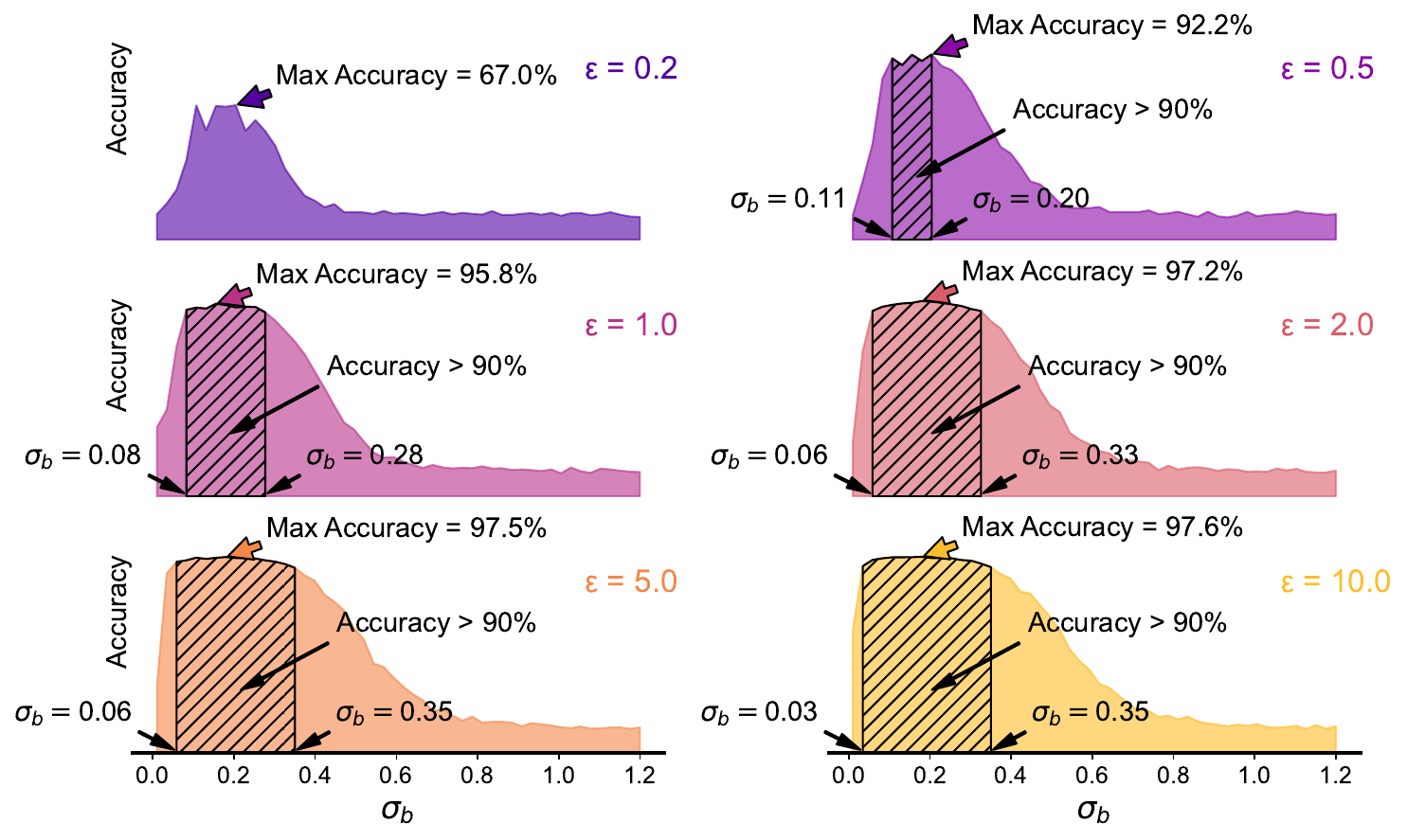}
    \caption{\textcolor{black}{Privacy-performance trade-off in differential privacy-hyperdimensional computing: impact of encoding on accuracy and privacy budget.}}
    \label{acc_train_heatmap}
\end{figure}

\textcolor{black}{Subsequently, we analyzed our dataset to determine the optimal \({\sigma}^{*}_{\text{b}}\) for our application. The core objective of our research was to find the best compromise between robust privacy protection and high operational efficiency. Figure~\ref{acc_train_heatmap} illustrates how the DP-HD model performs under different \(\epsilon\) levels with both exclusive and inclusive encoding strategies. Our efforts centered around finding a \({\sigma}^{*}_{\text{b}}\) that achieves high accuracy, demonstrating superior performance, while preserving a low \(\epsilon\), which reflects stronger privacy. In previous studies of HD, \({\sigma}_{\text{b}}\) is fixed at one, which we refer to as BaselineHD, the accuracy remains below 15\% across all \(\epsilon\) values, underscoring the significance of adapting \({\sigma}_{\text{b}}\) to balance privacy and performance effectively. In DP-HD, we selected \({\sigma}^{*}_{\text{b}}\), optimizing both accuracy and privacy. At this configuration, the model achieves the highest accuracy of 97.6\% with the least amount of noise, corresponding to the highest \(\epsilon\) of 10. As \(\epsilon\) decreases, privacy increases, but at the expense of accuracy. For instance, when \(\epsilon\) falls to 0.2, accuracy declines below 70\%. Conversely, when \(\epsilon\) exceeds 0.5, accuracy remains above 90\%. It is clear that for any level of privacy, as determined by choosing a specific value of \(\epsilon\), the \({\sigma}^{*}_{\text{b}}\) will be changed. This happens since there is a randomness in generating the random basis vectors that effect the model's performance. However, it is apparent for all privacy \(\epsilon\) values, the \({\sigma}^{*}_{\text{b}}\) is close to 0.2. Therefore we chose 0.2 as \({\sigma}^{*}_{\text{b}}\) to assign our dataset.}

\textcolor{black}{The uncertainty in DP-HD's output can be assessed by examining how much accuracy changes at a given privacy level when \(\sigma_{\text{b}}\) deviates from the optimal value, \(\sigma^{*}_{\text{b}}\). Specifically, we identify the range of \(\sigma_{\text{b}}\) values around \(\sigma^{*}_{\text{b}}\) that still allow accuracy to remain above 90\%, as illustrated in Figure~\ref{acc_train_heatmap}. For instance, at \(\epsilon = 10\), selecting \(\sigma_{\text{b}}\) between 0.03 and 0.35 ensures accuracy exceeds 90\%, which corresponds to a substantial 160\% variance from \(\sigma^{*}_{\text{b}} = 0.2\). This demonstrates that the DP-HD framework maintains a stable balance between accuracy and privacy even with a large range of \(\sigma_{\text{b}}\) deviations.}

\textcolor{black}{As \(\epsilon\) decreases, indicating a stronger privacy requirement, the range of acceptable \(\sigma_{\text{b}}\) values narrows slightly but remains robust. For example, at \(\epsilon = 5\) and \(\epsilon = 2\), \(\sigma_{\text{b}}\) can still vary between 0.06 and 0.33, offering 135\% flexibility around \(\sigma^{*}_{\text{b}}\) while sustaining accuracy above 90\%. At \(\epsilon = 1\), this range narrows further to 0.08–0.28, and at the high privacy level of \(\epsilon = 0.5\), \(\sigma_{\text{b}}\) can vary between 0.11 and 0.2, maintaining 45\% flexibility in \(\sigma_{\text{b}}\) selection. These findings indicate that DP-HD’s output uncertainty is minimal; by choosing an appropriate range for \(\sigma_{\text{b}}\), high accuracy is maintained across various privacy levels, underscoring the framework's reliability in balancing accuracy and privacy.}

 \begin{figure}
    \centering
    \includegraphics[width=\textwidth]{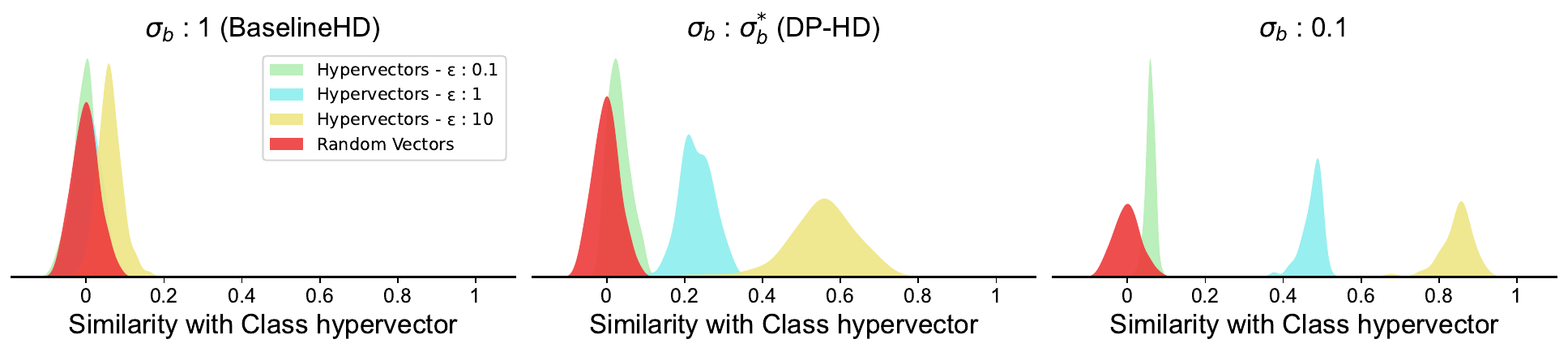}
    \caption{Effect of differential privacy on hyperdimensional computing memorization.}
    \label{mem_sim}
\end{figure}

 \begin{figure}
    \centering
    \begin{subfigure}[b]{0.49\textwidth}
        \centering
        \includegraphics[width=\textwidth]{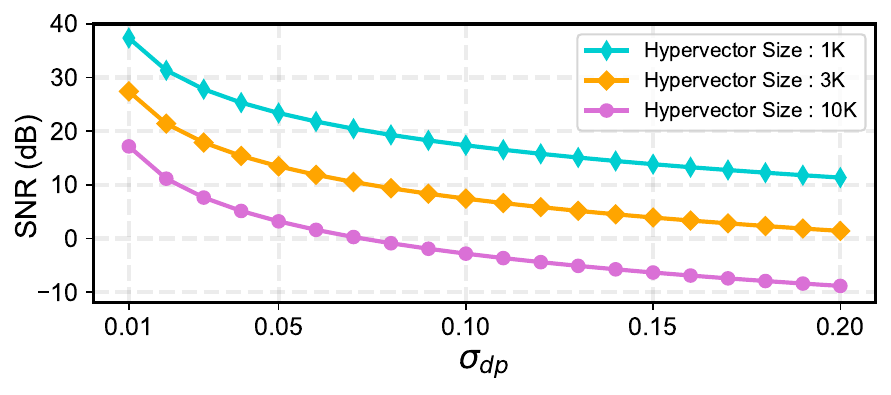}
        \caption{}
        \label{fig:snr_sigma}
    \end{subfigure}
    \hfill
    \begin{subfigure}[b]{0.49\textwidth}
        \centering
        \includegraphics[width=\textwidth]{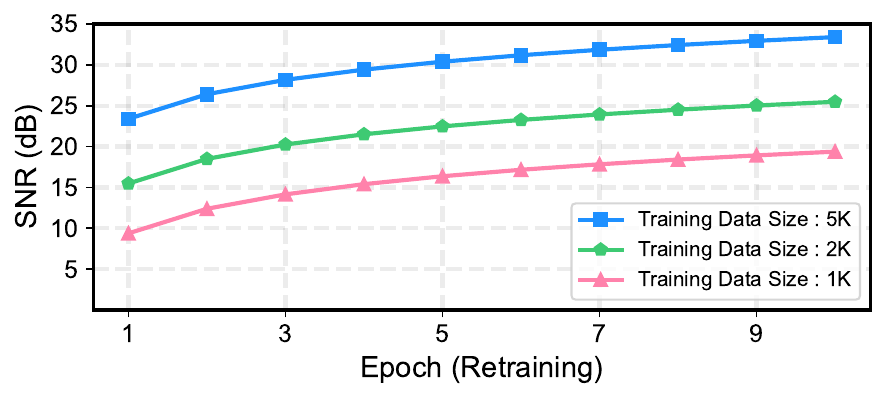}
        \caption{}
        \label{fig:snr_epoch}
    \end{subfigure}
    \caption{Differential privacy-hyperdimensional computing capability in control privacy and performance at the same time, (a) differential privacy-hyperdimensional computing capability in memorizing and separating the Signal and noise over different noise levels, (b) differential privacy-hyperdimensional computing capability in memorizing and separating the Signal and noise over different training epochs.}
    \label{fig:snr}
\end{figure}

After finding the best encoding strategy, it is important to choose a privacy level that does not drop the model's performance dramatically. The privacy level is chosen by choosing the \(\epsilon\) value. A smaller \(\epsilon\) value indicates the introduction of greater noise levels. This added noise diminishes the ability of class hypervectors within the DP-HD model to retain training sample information. Although increased noise enhances privacy, it comes at the cost of reduced DP-HD performance. Figure~\ref{mem_sim} displays how varying noise amounts affect DP-HD memorization. For instance, with \({\sigma}_{\text{b}}\) at 0.1, the similarity between randomly generated vectors and noisy class hypervectors approaches zero, indicating that the DP-HD model fails to retain any information from these random vectors. At a higher \(\epsilon\) value of 10, the noise level is lower, thus maintaining a high similarity between hypervectors and class hypervectors because the noise and signal distributions remain distinct. Reducing \(\epsilon\) increases privacy but at the expense of losing vital training sample information, leading to a decrease in similarity. For example, at an \(\epsilon\) of 1, the central similarity distribution shifts to around 0.5. A substantial increase in noise, such as when \(\epsilon\) is set to 0.1, drives the DP-HD model to store predominantly noise information, pushing similarity values close to those of random vectors. Additionally, \({\sigma}_{\text{b}}\) plays a critical role in DP-HD memorization. In BaselineHD, with a \({\sigma}_{\text{b}}\) of 1, the similarity of hypervectors to class hypervectors remains unaffected by \(\epsilon\) variations and consistently hovers near zero, explaining the consistently low accuracy of BaselineHD regardless of the noise level. Selecting an optimal \({\sigma}^{*}_{\text{b}}\) reduces similarity, yet the distributions of hypervectors and random vectors remain distinct. For instance, even with an \(\epsilon\) of 10, similarity decreases from approximately 0.85 to 0.55, demonstrating that while an inclusive encoding strategy may increase similarity, it does not necessarily translate to higher accuracy.

Moreover, the DP-HD explainability property, let us to predict the impact of the noise on the model's performance by measuring the DP-HD's ability to discriminate the signal and noise. Figure~\ref{fig:snr_sigma} illustrates the influence of noise on the ability of the DP-HD model to distinguish between signal and noise during training, which is critical for accurately labeling queries in the inference phase. As the noise level increases, the SNR diminishes. Conversely, reducing the size of the hypervector leads to an increase in SNR. For instance, an SNR of 17.17dB is observed with a hypervector size of 10K and a \({\sigma}_{\text{dp}}\) of 0.01. Escalating the noise level to 0.2 results in a decrease of SNR to -8.85dB. Reducing the hypervector size to 1K, however, raises the SNR to 37.37dB. Additionally, enhancements in training—such as increasing the training dataset size—are shown to elevate the SNR, as depicted in Figure~\ref{fig:snr_epoch}. With a training size of 1K and after one retraining epoch, the SNR is 9.37dB. Following five and ten retraining epochs, SNR rises to 16.36dB and 19.37dB, respectively. Larger training sets further elevate SNR to 25.47dB and 33.39dB for sizes of 2K and 5K, respectively. These findings demonstrate the ability to modulate the added noise in the DP-HD model, thereby balancing privacy with performance. They also confirm that the impact of privacy measures on accuracy can be assessed prior to noise integration, unlike other ML models.

\subsubsection{Benchmark Comparison}
\textcolor{black}{In this study, we benchmarked our newly developed DP-HD model against established models in the field of image classification. This comparison includes an evaluation alongside models such as ResNet50, GoogLeNet, AlexNet, DenseNet201, and EfficientNet B2. To implement DP-SGD, we utilized Opacus, a freely available open-source library designed for PyTorch that facilitates training deep learning models with DP mechanisms. Table~\ref{table:bench_table_acc} details the accuracies post-DP application, revealing that DP-HD sustains a slight accuracy reduction compared to significant drops in other models. With a stringent privacy level, represented by an \(\epsilon\) value of 0.6, DP-HD maintains an accuracy of 91.04\%, whereas other models, securing using DP-SGD, plummet below 43\%, showcasing DP-HD's robustness against high noise levels as it distributes memory evenly across all dimensions of the hyperspace. As \(\epsilon\) increases to 0.8, reducing the noise, DP-HD's accuracy climbs to 93.34\%, significantly surpassing other models which remain under 54\%. This performance gap expands further when \(\epsilon\) is adjusted to 1, with DP-HD reaching an accuracy of 94.43\%, and the closest competitor, DenseNet201, achieves only 69.13\%, still 25.3\% lower than DP-HD. As \(\epsilon\) increases further to 2, the accuracy of DP-HD is 96.30\%, while the other models' accuracies—75.22\% for ResNet50, 82.48\% for GoogLeNet, 58.20\% for AlexNet, 77.70\% for DenseNet201, and 57.20\% for EfficientNet B2—illustrate a substantial disparity, with GoogLeNet being the closest yet still 13.82\% less accurate than DP-HD. Even at an \(\epsilon\) of 3, where privacy levels are lower and noise reduction is significant, DP-HD's accuracy remains high at 96.66\%, outperforming ResNet50, GoogLeNet, AlexNet, DenseNet201, and EfficientNet B2, which exhibit accuracies of 81.18\%, 83.73\%, 56.27\%, 86.89\%, and 56.46\% respectively. This demonstrates DP-HD's superior capability to balance privacy and performance compared to other popular ML models in the field.}

\begin{table}[h!]
\centering
\color{black}

\caption{Defect detection accuracy across different privacy budget values.}
\begin{tabular}{c c c c c c c}

\hline
Privacy Mechanism & Model  & $\epsilon$ = 0.6 & $\epsilon$ = 0.8 & $\epsilon$ = 1 & $\epsilon$ = 2 & $\epsilon$ = 3 \\
\hline
\textbf{DP-HD} & \textbf{HD}  & \textbf{91.04\%} & \textbf{93.34\%} & \textbf{94.43\%} & \textbf{96.30\%} & \textbf{96.66\%} \\
DP-SGD & ResNet50   & 30.93\% & 45.28\% & 52.30\% & 75.22\% & 81.18\% \\
DP-SGD & GoogLeNet  & 12.48\% & 12.98\% & 23.85\% & 82.48\% & 83.73\% \\
DP-SGD & AlexNet   & 42.98\% & 53.11\% & 55.78\% & 58.20\% & 56.27\% \\
DP-SGD & DenseNet201  & 41.12\% & 51.99\% & 69.13\% & 77.70\% & 86.89\% \\
 DP-SGD & EfficientNet B2  & 32.24\% & 35.22\% & 40.81\% & 57.20\% & 56.46\% \\
RDP & ResNet50 & 31.55\% & 36.71\% & 61.37\% & 75.78\% & 88.94\% \\
RDP & GoogLeNet & 12.48\% & 38.45\% & 40.99\% & 73.42\% & 79.69\% \\
RDP & AlexNet & 34.66\% & 57.70\% & 52.92\% & 55.03\% & 55.59\% \\
RDP & DenseNet201 & 29.19\% & 17.02\% & 68.26\% & 66.15\% & 83.17\% \\
RDP & EfficientNet B2 & 29.69\% & 44.66\% & 47.08\% & 61.24\% & 63.98\% \\
 \hline
\end{tabular}
\label{table:bench_table_acc}
\end{table}

\textcolor{black}{For the comparison of DP-HD with RDP, Table~\ref{table:bench_table_acc} clearly demonstrates the superior performance of DP-HD across all privacy budget values. At the lowest privacy budget (\(\epsilon = 0.6\)), DP-HD achieves an accuracy of 91.04\%, while the highest-performing RDP-based model, AlexNet, reaches only 34.66\%, resulting in a gap of 56.38\%. This substantial difference highlights DP-HD's robustness in maintaining high accuracy even under strict privacy constraints. As \(\epsilon\) increases to 0.8, DP-HD's accuracy improves to 93.34\%, whereas AlexNet, the best RDP-based model, only reaches 57.70\%, maintaining a difference of 35.64\%. At \(\epsilon = 1\), DP-HD achieves 94.43\%, significantly outperforming DenseNet201 (68.26\%) and other RDP-based models, such as ResNet50 (61.37\%) and AlexNet (52.92\%), with at least a 26.17\% margin over the best competitor. At higher privacy budgets, such as \(\epsilon = 2\), DP-HD maintains its dominance with an accuracy of 96.30\%, compared to ResNet50's 75.78\%, DenseNet201's 66.15\%, and GoogLeNet's 73.42\%. The gap between DP-HD and the top-performing RDP-based model remains significant, at 20.52\%. Finally, at \(\epsilon = 3\), DP-HD achieves a remarkable accuracy of 96.66\%, surpassing ResNet50 (88.94\%) and DenseNet201 (83.17\%) by 7.72\% and 13.49\%, respectively. This consistent outperformance across all privacy budgets emphasizes DP-HD's ability to balance high accuracy and strong privacy guarantees more effectively than RDP-based implementations.}

\begin{figure}
    \centering
    \begin{subfigure}[b]{0.49\textwidth}
        \centering
        \includegraphics[width=\textwidth]{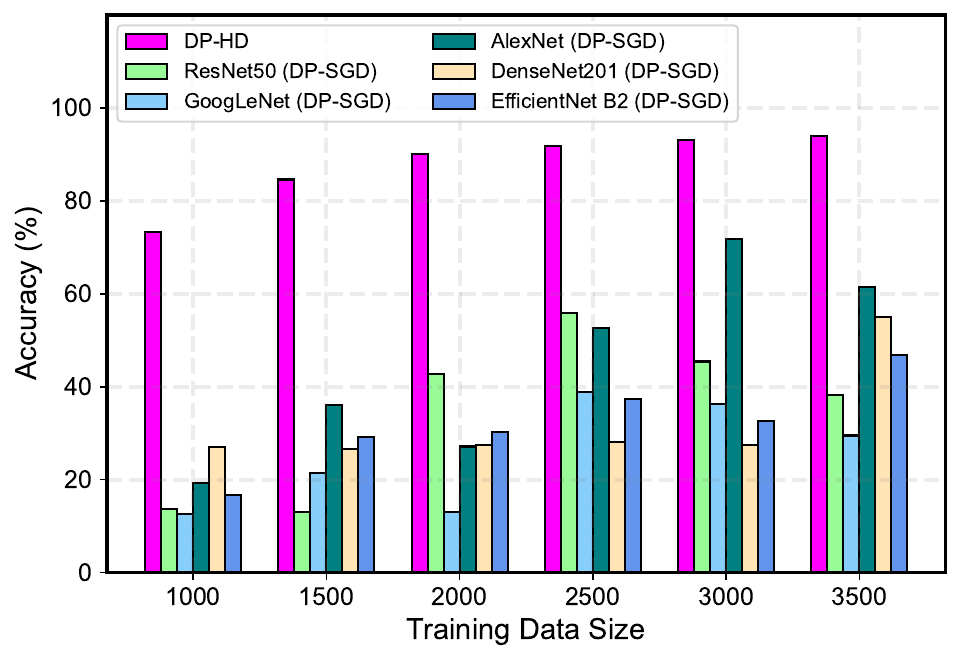}
        \caption{}
        \label{fig:acc_tr_dpsgd}
    \end{subfigure}
    \hfill
    \begin{subfigure}[b]{0.49\textwidth}
        \centering
        \includegraphics[width=\textwidth]{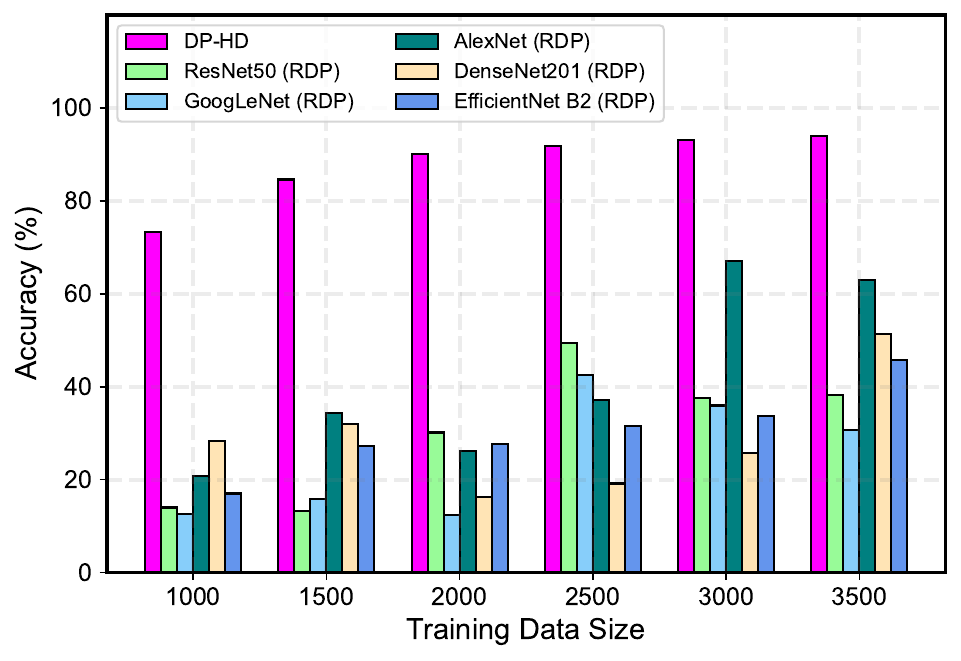}
        \caption{}
        \label{fig:acc_tr_rdp}
    \end{subfigure}
    \caption{\textcolor{black}{Defect detection accuracy across varying training sample sizes, comparing the proposed DP-HD method with (a) DP-SGD and (b) RDP-based benchmarks.}}
    \label{fig:acc_tr}
\end{figure}

\textcolor{black}{When evaluating DP-HD against both DP-SGD and RDP methods across varying training set sizes, with the privacy budget fixed at \(\epsilon = 1\), DP-HD continues to showcase superior performance. As seen in Figure~\ref{fig:acc_tr_dpsgd}, and \ref{fig:acc_tr_rdp}, DP-HD achieves an accuracy of 73.26\% at the smallest training size of 1,000 samples, significantly outperforming ResNet50 with DP-SGD (13.66\%) and RDP (14.00\%). This trend persists as the training size increases, with DP-HD maintaining its lead across all benchmarks. At a training size of 3,500 samples, DP-HD reaches 94.02\% accuracy, while the closest competitor, AlexNet using DP-SGD, achieves 61.37\%. RDP-based AlexNet follows with 62.88\%, still trailing by over 31.14\%.}

\textcolor{black}{The robustness of DP-HD becomes even more apparent when comparing its consistency across different training set sizes. Models using DP-SGD, such as ResNet50 and GoogLeNet, exhibit fluctuating accuracies, often struggling to surpass 40\% at smaller training sizes. Similarly, RDP-based models, while slightly more stable, still fail to match the consistent improvements of DP-HD as the training size grows. These results highlight DP-HD's ability to leverage its hyperdimensional representation for enhanced accuracy and stability, solidifying its position as a robust solution for privacy-preserving ML tasks.}

\subsection{Inference Privacy} 

\begin{figure}
    \centering
    \includegraphics[width=\textwidth]{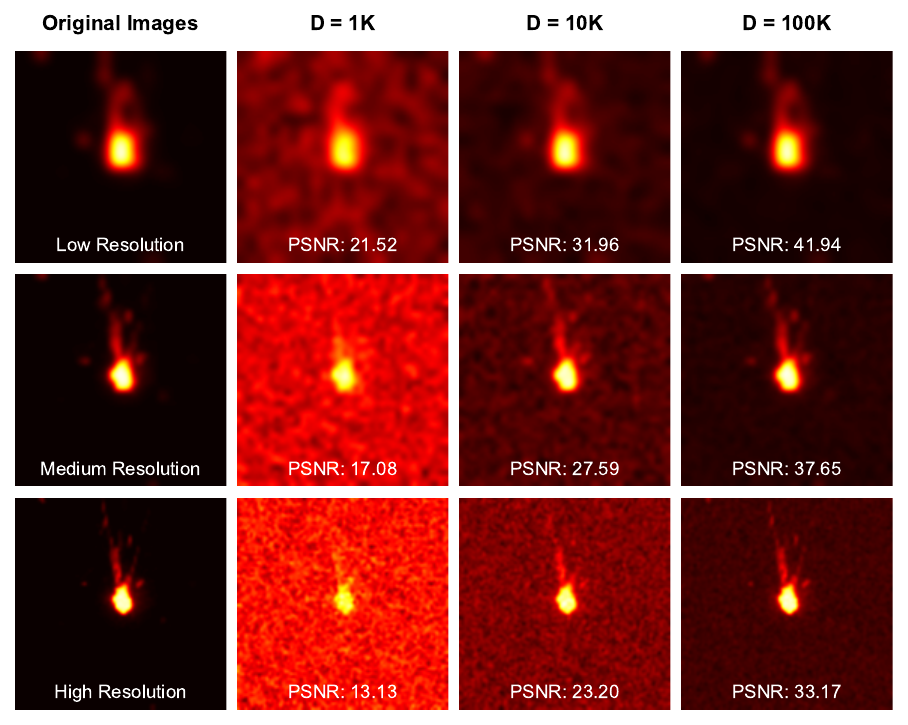}
        \caption{Decoding optical images of laser powder bed fusion process to capture the confidential information.}
    \label{decod_eg}
\end{figure}

The HD model exhibits vulnerabilities during the inference phase, particularly when an encoded query is transmitted, as attackers can readily decode it. The encoding process is defined as \(h_{i} = \cos(\Vec{F} \cdot \Vec{B}_{i})\), allowing for the original feature vector to be reconstructed by Equation~\eqref{eq:inference decode}. Figure~\ref{decod_eg} showcases the retrieval of optical images at low, medium, and high resolutions. It is observed that increasing image resolution necessitates larger hypervector sizes to preserve complete information. For low-resolution images, a 1K-sized hypervector suffices, yielding a reconstruction with a Peak Signal-to-Noise Ratio (PSNR) of 21.52. As the hypervector size expands, it captures more data from the feature vector, enhancing the PSNR to 31.96 and 41.94 for dimensions of 10K and 100K, respectively. In the case of medium-resolution images, a hypervector size of 1K results in a dissimilar reconstruction with a PSNR of 17.08; however, increasing \(D\) to 100K improves the PSNR to 37.65, preserving more pixel values within the hyperspace. For high-resolution images, substantial hypervector dimensions are required. For instance, with \(D\) settings of 1K and 10K, reconstructing a high-resolution image accurately is challenging. Conversely, a \(D\) of 100K facilitates a reconstruction with a PSNR of 33.17, indicating that larger \(D\) values are essential for effectively capturing all pixel information in high-resolution scenarios.

\begin{figure}
    \centering
    \includegraphics[width=0.85\textwidth]{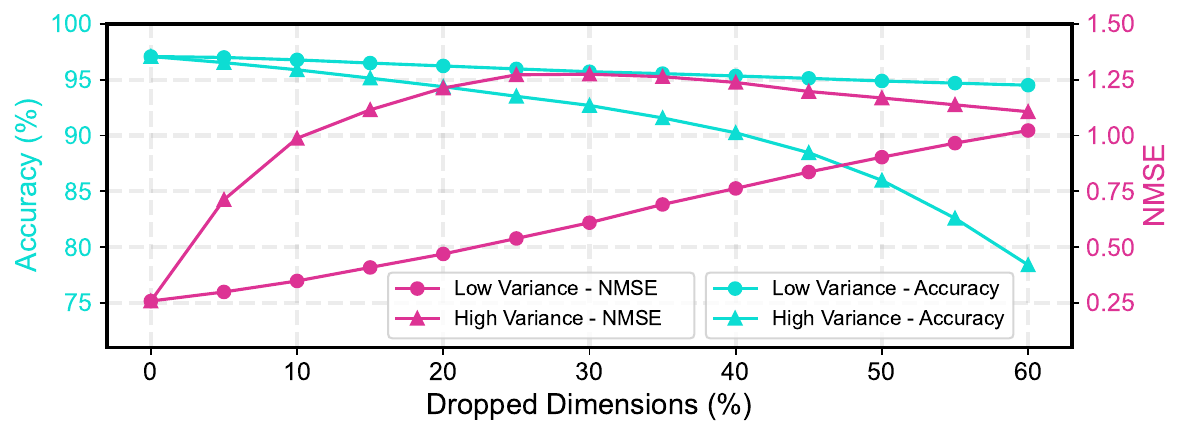}
    \caption{Accuracy and normalized mean square error after dropping low variance dimensions to secure query hypervectors.}
    \label{drop_acc}
\end{figure}

In AM, visual imaging sensors consistently capture images and videos that follow specific structures and patterns. Even with nonlinear encoding approaches, ML models can readily extract sensitive information from hypervectors. Consequently, it becomes imperative to implement a privacy mechanism during the inference phase of the HD to protect this data. Reducing the size of the hypervector generally complicates the task for attackers trying to decode it and extract feature information. In cases of encoded queries, dimensions with low variance carry minimal meaningful information, as they do not vary significantly across different classes. By setting these low-variance dimensions to zero, not only is it difficult for attackers to discern information about the images, but the accuracy of the HD model remains largely unaffected. Figure~\ref{drop_acc} demonstrates the effects of removing low and high variance dimensions from encoded queries on both HD model performance in classification tasks and the ability of attackers to reconstruct raw data. Eliminating low-variance dimensions does not substantially reduce accuracy. For instance, discarding 60\% of these dimensions results in a minor accuracy decrease from 97.08\% to 94.51\%, marking a 2.57\% drop in HD performance. In contrast, the Normalized Mean Square Error (NMSE) increases significantly from 0.25 to 1.02, a 308\% rise, indicating a reduced risk of data recovery by attackers. Conversely, removing high-variance dimensions, which contain crucial information, leads to a sharp decline in model accuracy. Dropping 60\% of these dimensions results in an NMSE of 1.10, comparable to when low-variance dimensions are removed. However, accuracy plummets to 78.40\%, reflecting an 18.67\% performance decrease in the HD model.

\begin{figure}
    \centering
    \includegraphics[width=0.95\textwidth]{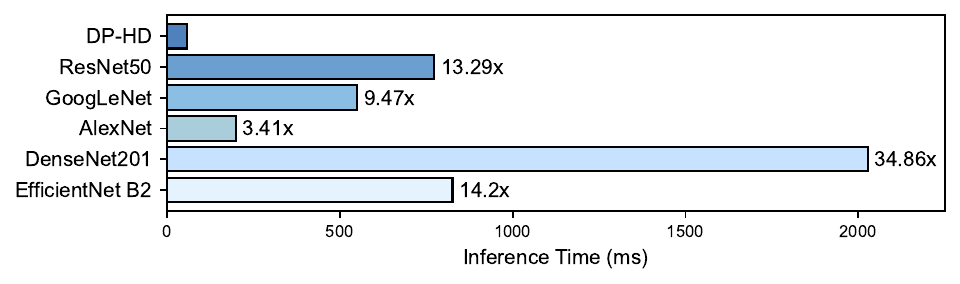}
    \caption{Inference time for defect detection of 1610 query samples.}
    \label{fig:inf_time_bench}
\end{figure}

An ML model employed for in-situ sensing in AM necessitates a brief inference time to be effective. Figure~\ref{fig:inf_time_bench} demonstrates the inference times for DP-HD compared to other ML models when predicting labels for 1610 query samples. The inference times are as follows: DP-HD at 58.15ms, ResNet50 at 772.78ms, GoogLeNet at 550.51ms, AlexNet at 198.09ms, DenseNet201 at 2027.44ms, and EfficientNet B2 at 825.94ms. DP-HD's markedly lower inference time is attributed to its computational simplicity, requiring only the calculation of the dot product between hypervectors. AlexNet has the closest inference time to DP-HD, at approximately 3.41 times that of DP-HD, while DenseNet201 has the longest, at 34.86 times DP-HD's time. The higher inference times for the other models, except DP-HD, can be attributed to their complex architectures which involve multiple layers and operations such as convolutions and pooling, making them computationally intensive. These complexities underscore DP-HD’s advantage as the fastest option for in-situ sensing in AM, providing swift and efficient processing that is crucial for real-time applications. Its speed ensures minimal latency, making DP-HD particularly suitable for AM environments where quick decision-making is critical.

\section{\textcolor{black}{Discussion}}
\label{sec:Discussion}
\textcolor{black}{In this paper, we introduced the DP-HD framework to secure sensitive information from in-situ sensing in AM environments. By combining DP with HD, DP-HD enhances privacy protection across both training and inference phases while achieving a balance between accuracy and privacy. This balance is quantified using an SNR, which measures the contribution of training samples to HD performance relative to the noise from DP. Below, we discuss the core components of our framework, as well as its strengths and limitations.}

\textcolor{black}{Our DP-HD framework leverages HD encoding to map feature vectors into a high-dimensional space, using a mathematically grounded approach that relies on principles inspired by Bochner’s theorem \cite{bochner1933monotone}. According to Bochner’s theorem, any continuous, shift-invariant kernel can be represented through a Fourier transform. Using Random Fourier Features (RFF) \cite{rahimi2007random}, our framework approximates this mapping, which allows the model to generalize across distributions effectively. Rather than depending on specific HD parameters or sample size, the optimal variance for RFF is derived directly from the underlying data distribution \cite{sutherland2015error}, allowing our model to adapt well to datasets with broad and representative distributions. Consequently, this encoding technique ensures that conclusions drawn from one dataset are not constrained to that dataset but are applicable to other datasets with similar distributions.}

\textcolor{black}{The importance of dataset distributional richness also plays a significant role in balancing accuracy and privacy. To achieve this, we rely on datasets that span a wide feature space, ensuring that the RFF variance captures essential data patterns. This distributional breadth is critical for extending our model’s utility across various datasets, emphasizing that conclusions are grounded in distribution characteristics rather than specific data points. As such, the robustness of our framework depends more on the diversity within a dataset’s distribution than on any individual samples, which enhances the model’s applicability to new datasets in different settings.}

\textcolor{black}{However, certain limitations arise when extending the DP-HD framework to larger datasets or real-world applications. While the RFF approach is computationally efficient, it may degrade when dealing with complex or non-stationary data distributions. In practice, real-world data may not fully align with assumptions of shift-invariance or smoothness required for accurate kernel approximations \cite{yang2012nystrom}. Additionally, the variance chosen for RFF is critical. Although our method optimizes this variance based on the data’s second moment, real-world datasets with skewed or multimodal distributions might require further refinement to capture the nuances in data patterns effectively.}

\textcolor{black}{The choice of exclusive versus inclusive encoding in HD also impacts the balance between accuracy and privacy. As our results indicate, encoding choices should adapt based on the distribution within each class and the distance between class centers to maintain high accuracy and privacy. Importantly, our findings reveal that the optimal SD of random basis vectors remains stable across different HD configurations, as long as the dataset distribution is adequately represented. This stability further reinforces that our framework’s reliability does not depend on specific HD model parameters or sample sizes, enhancing its robustness.}

\textcolor{black}{Additionally, the only hyperparameter requiring manual selection in the DP-HD framework is the size of the hypervectors, which must be sufficiently large to capture all essential information from the feature vectors. For our experimental setup, we selected a hypervector size of 1000, as this was the minimum size required to preserve relevant information effectively. To validate this choice, we tested the framework with higher hypervector sizes and found that while larger sizes resulted in marginal accuracy improvements, the gains were negligible and did not justify the additional computational cost. This demonstrates the robustness of HD-based models, which maintain stable performance as long as the hypervector size exceeds a certain threshold. In the case of DP-HD, this threshold was determined to be 1000, ensuring reliable performance without extensive hyperparameter tuning.}

\textcolor{black}{We also observe that DP-HD’s uncertainty is minimal. By selecting an appropriate range of standard deviations for random basis vectors, we maintain high accuracy across privacy levels. Furthermore, in terms of inference privacy, selectively dropping dimensions from query hypervectors effectively protects query-specific information, thus misleading potential adversaries attempting to reconstruct the original data.}

\textcolor{black}{Finally, we benchmarked DP-HD against state-of-the-art models for both training and inference privacy. Our results indicate that DP-HD consistently outperforms these models, offering higher levels of privacy and accuracy along with faster inference times, which is essential for real-time applications such as in-situ sensing in AM.}

\textcolor{black}{For future work, the proposed DP-HD framework can be extended to other domains that require a balance between privacy and accuracy in ML models trained on sensitive data. While this study focuses on AM, its utility is particularly relevant for privacy-sensitive industrial environments such as aerospace, automotive, and energy production. For instance, real-time defect detection in these sectors often involves proprietary designs and sensitive operational data, where privacy breaches could have financial and regulatory consequences. The DP-HD framework provides a scalable solution to address these challenges while maintaining high accuracy. DP-HD can also be applied in sectors such as healthcare, financial services, and government data management. These domains frequently involve training ML models on confidential information, where ensuring data privacy is crucial while maintaining high model performance.}

\textcolor{black}{In healthcare, ML models are often trained on sensitive patient data, including medical histories and diagnostic records, which are vulnerable to privacy breaches. Financial services use ML models for tasks like fraud detection and risk assessment, which rely on sensitive customer information, such as transaction records and credit history. Government data management can involve ML models analyzing citizen data, including personal identification and behavioral records, which are susceptible to misuse if privacy is compromised. Applying the DP-HD framework in these areas can help protect confidential information while ensuring that model accuracy remains effective for these critical applications.}

\section{Conclusion and Future Works}
\label{sec:Conclusions}

This paper introduced the Differential Privacy-Hyperdimensional Computing (DP-HD) framework as an innovative solution for addressing privacy concerns in Additive Manufacturing (AM). By integrating Differential Privacy (DP) mechanisms within the Hyperdimensional Computing (HD) model, the DP-HD framework ensures the protection of sensitive data during training and inference while maintaining high model performance. The key advantage of DP-HD is its explainability, enabling the prediction of noise impact on model accuracy, which is critical for in-situ sensing applications in AM. Our experiments demonstrated that DP-HD preserves privacy without significantly compromising accuracy, showcasing its robustness and suitability for real-time monitoring in AM. The ability to balance privacy and performance through explainable insights sets DP-HD apart, making it a valuable tool to improve manufacturing precision while safeguarding data privacy. While the DP-HD framework demonstrates significant promise, certain aspects warrant further exploration to enhance its applicability. Currently, finding the optimal encoding strategy is based on the selection of training and validation subsets. Future work focuses on developing mathematical methods to automatically determine the best Standard Deviation (SD) for generating random basis vectors based on data distribution. Moreover, deploying DP-HD in collaboration with industry partners can validate its scalability and adaptability for practical industrial applications, such as real time defect detection in aerospace and automotive manufacturing, where privacy concerns often intersect with stringent performance requirements. Additionally, due to DPHD’s success in privacy preserving techniques, the framework can be integrated within a federated learning structure to leverage distributed data processing while maintaining robust privacy measures. Beyond AM, the DP-HD framework holds considerable potential in other domains requiring secure and efficient machine learning applications. Future work will explore its applicability in areas such as healthcare, financial services, and government data management. In these fields, the ability to maintain a delicate balance between data privacy and model performance is crucial for safeguarding sensitive information while delivering reliable, high quality predictions.

\section*{Acknowledgment}
This work was supported by the National Science Foundation, United States [grant numbers 2127780, 2312517, 2434519]; the Semiconductor Research Corporation (SRC), United States; the Office of Naval Research, United States [grant numbers N00014-21-1-2225, N00014-22-1-2067]; the Air Force Office of Scientific Research, United States [grant number FA9550-22-1-0253]; UConn Startup Funding, and generous gifts from Xilinx and Cisco. The authors gratefully acknowledge the valuable contributions of the National Institute of Standards and Technology (NIST, United States), particularly Dr. Brandon Lane, who provided data for this research.

\section*{Supplementary Materials}
\textcolor{black}{The code used in this study is openly available for reproducibility and can be accessed at the following link: \href{https://github.com/FardinJalilPiran/Differential-Privacy-Hyperdimensional-Computing.git}{https://github.com/FardinJalilPiran/Differential-Privacy-Hyperdimensional-Computing.git}}.

\bibliographystyle{unsrt}  
\bibliography{references}

\end{document}